\documentclass{article}

    \PassOptionsToPackage{numbers, compress}{natbib}
 \usepackage[preprint]{neurips_2026}

\usepackage[utf8]{inputenc} % allow utf-8 input
\usepackage[T1]{fontenc}    % use 8-bit T1 fonts
\usepackage{hyperref}       % hyperlinks
\usepackage{url}            % simple URL typesetting
\usepackage{booktabs}       % professional-quality tables
\usepackage{amsfonts}       % blackboard math symbols
\usepackage{nicefrac}       % compact symbols for 1/2, etc.
\usepackage{microtype}      % microtypography
\usepackage{xcolor}         % colors

\usepackage{amsmath}
\usepackage{algorithm}
\usepackage{algorithmic}
\usepackage{multirow}
\usepackage{makecell}
\usepackage{booktabs}
\usepackage{graphicx}
\usepackage[normalem]{ulem} % For underlining
\usepackage{array}
\usepackage[table]{xcolor}
\definecolor{bestgreen}{RGB}{226,239,218}
\usepackage{amssymb}
\usepackage{pifont}

\newcommand{\xmark}{\ding{55}}

\title{FM-ChangeNet: Learning Change through Pathwise Feature Transport}

\author{%
  Roie Kazoom \\
  Google Research \\
  \texttt{kazoomroie@google.com}
  \And
  George Leifman \\
  Google Research \\
  \texttt{gleifman@google.com}
  \And
  Genady Beryozkin \\
  Google Research \\
  \texttt{genady@google.com}
}

\begin{document}

\maketitle

\vspace{-10pt}

\begin{figure*}[!h]
  \centering
  \includegraphics[width=0.45\textwidth]{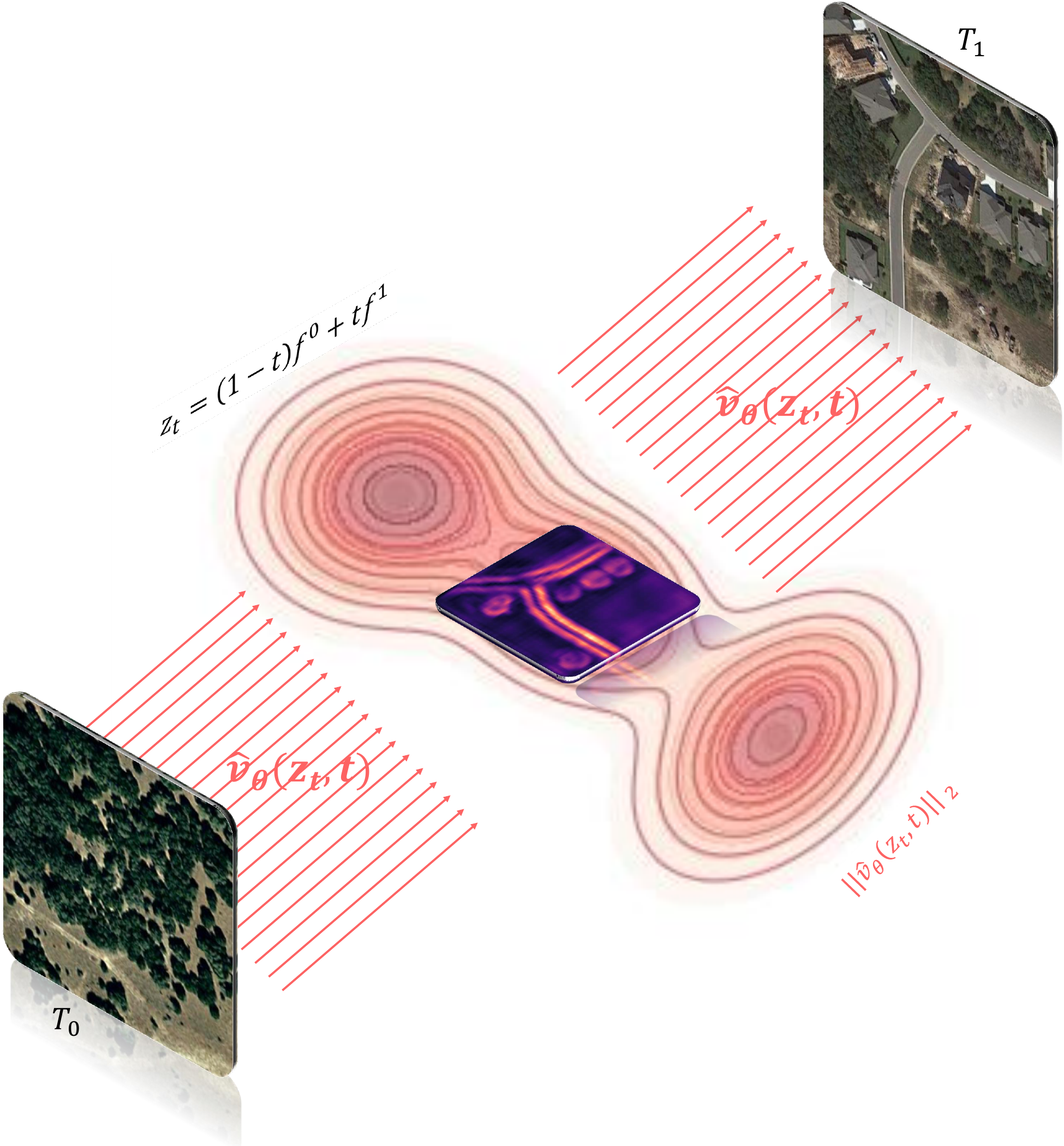}
\caption{\textbf{Conceptual illustration of flow matching for change detection.}
    Given a bi-temporal image pair $(T_0, T_1)$, the model defines an intermediate latent state
    $z_t = (1-t)f^0 + t f^1$ along a continuous feature-space interpolation between the encoded
    pre and post-temporal representations. A time-conditioned velocity field
    $\hat{v}_{\theta}(z_t,t)$ characterizes the local transport dynamics from $T_0$ toward $T_1$,
    while its magnitude $\|\hat{v}_{\theta}(z_t,t)\|_2$ provides a spatially localized cue for
    change.}
    \label{fig:flow_matching_concept}
\end{figure*}

\begin{abstract}
We present \textbf{FM-ChangeNet}, a \emph{pathwise-supervised} framework for change detection that reformulates bi-temporal reasoning as \emph{continuous transport in feature space} rather than static endpoint comparison. Given encoded pre and post-temporal representations, we construct intermediate latent states and learn a time-conditioned velocity field $\hat{v}_\theta(z_t,t)$ along the transformation trajectory. This pathwise formulation constrains the predictor over a continuum of intermediate states, providing a denser and less ambiguous supervision signal than conventional endpoint-only segmentation and enabling the model to capture temporal evolution explicitly. The learned velocity field is not only a transport mechanism but also an interpretable representation of change: its magnitude serves as a spatially localized change cue that helps distinguish true structural variation from nuisance effects such as illumination shifts and spatial misalignment. We develop a hierarchical multi-scale architecture with cross-temporal alignment, time-conditioned coarse-to-fine flow decoding, and a unified objective that couples flow supervision, trajectory consistency, spatial regularization, and segmentation loss. Experiments on remote sensing benchmarks show that the proposed framework produces more structured and robust change representations while achieving state-of-the-art performance.
\end{abstract}

\begin{figure}[!t]
  \centering
  \includegraphics[width=0.85\textwidth]{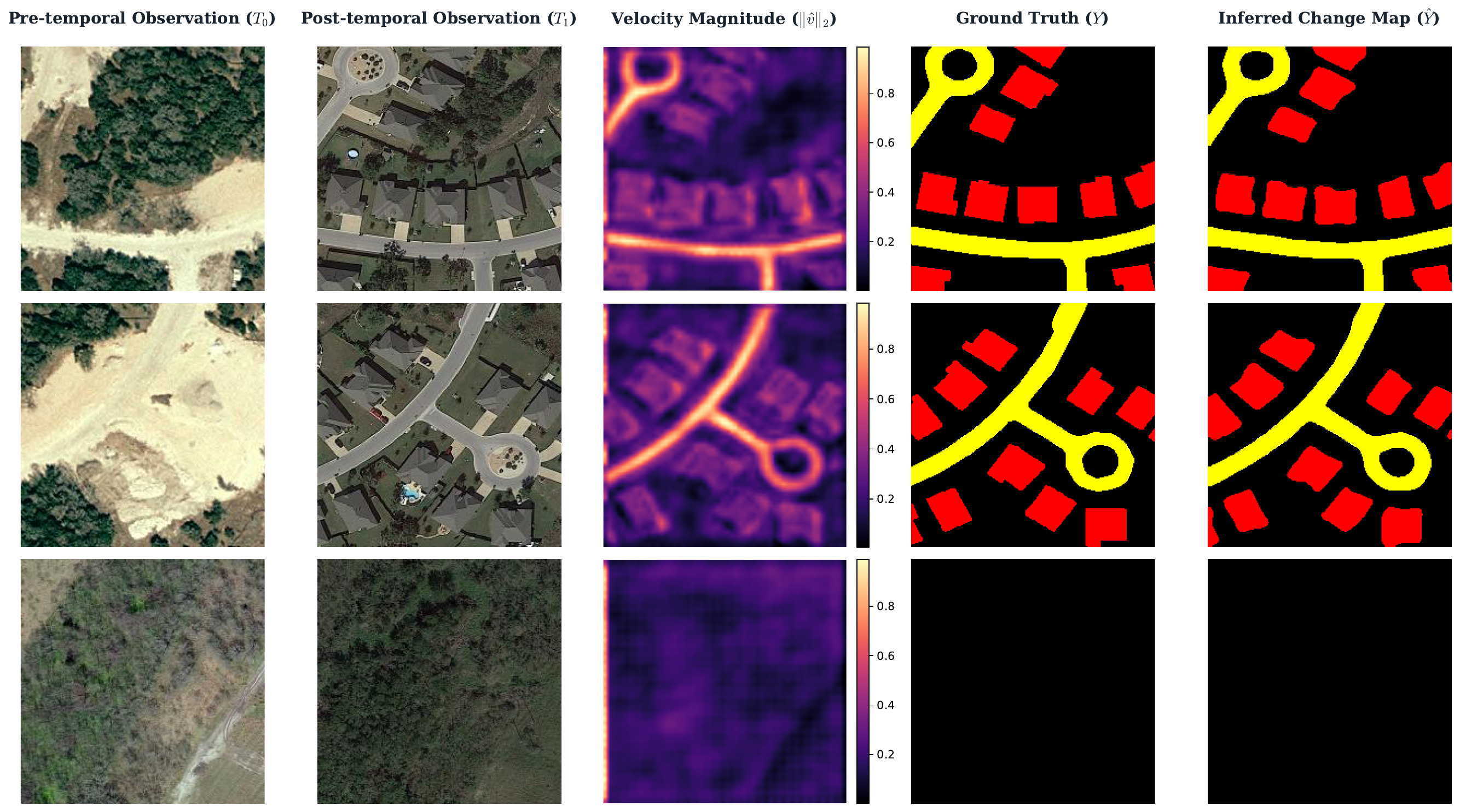}
\caption{\textbf{Change detection as feature-space transport.}
Given a bi-temporal pair $(T_0,T_1)$, the encoder extracts feature representations $f^0=E(T_0)$ and $f^1=E(T_1)$. We model their relationship through a continuous latent path $z_t=(1-t)f^0+t f^1$ and learn a time-conditioned velocity field $\hat{v}_\theta(z_t,t)$ using flow matching. The velocity magnitude $\|\hat{v}_\theta(z_t,t)\|_2$ provides a localized cue for change: unchanged regions require little transport, whereas structurally changed regions induce coherent high-magnitude responses. This transport-based representation aligns with the ground-truth mask $Y$ and supports accurate prediction of the final change map $\hat{Y}$.}
  \label{fig:teaser}
\end{figure}

\section{Introduction}

Change detection (CD) aims to identify meaningful surface variations between bi-temporal images captured over the same geographical area at different times. It is a fundamental problem in remote sensing, with applications in urban expansion monitoring \cite{chen2020levircd, van2021spacenet7}, disaster assessment \cite{zhu2017deep}, and environmental conservation. Reliably isolating changes remains challenging due to nuisance factors such as illumination variation, seasonal effects, and viewpoint shifts, particularly in high-resolution imagery \cite{chen2025deeplcd, zhang2020ifnet}. Moreover, detecting subtle or small-scale changes often leads to high false-alarm rates when relying on local spectral differences \cite{lei2023ussfc}. Modern CD methods are dominated by Siamese architectures, where bi-temporal inputs are processed independently and fused via concatenation, differencing, or attention mechanisms. Early works such as FC-Siam \cite{daudt2018fully} and IFNet \cite{zhang2020ifnet} established this paradigm, while more recent approaches incorporate Transformers \cite{chen2021bit, bandara2022changeformer} or dense multi-scale designs \cite{fang2021snunet} to improve representation quality. Despite their strong empirical performance, these methods share a fundamental limitation: they treat the two observations as static snapshots and infer change from implicit feature comparisons. As a result, supervision is imposed primarily at the endpoint through the final change mask, while the underlying \emph{temporal transition}, namely the process that transforms one state into the other, remains largely unconstrained. This limitation becomes especially pronounced under gradual structural evolution, cross-resolution discrepancies, and spatial misalignment, where endpoint differencing may confound true change with nuisance variation \cite{chen2023continuous, tian2022racdnet}. Recent generative approaches attempt to address this limitation by modeling the distribution of change features, for example using diffusion models \cite{Bandara_2025_WACV, tang2024changeanywhere}. While effective, such methods typically rely on stochastic sampling procedures and still do not explicitly characterize the transformation pathway between observations. Consequently, they improve the representation of change features, but do not directly supervise \emph{how} pre-temporal structure evolves into post-temporal structure. This raises a fundamental question: \textbf{can change detection be formulated as learning the transformation itself, rather than detecting differences only at the endpoint?} 

In this work, we answer this question by reframing change detection as a \emph{pathwise-supervised feature-space transport problem}. Inspired by Flow Matching (FM) \cite{lipman2022flow, liu2022flow, tong2023improving}, we model the relationship between bi-temporal representations as a continuous transformation governed by a time-conditioned velocity field. Rather than relying solely on an endpoint-fused representation for mask prediction, FM-ChangeNet is constrained to predict transport over intermediate latent states. The fused deepest-scale representation initializes cross-temporal alignment and coarse decoding, while the final prediction is conditioned on pathwise velocity estimates across the multi-scale trajectory. This reformulation changes the \emph{supervision structure} of CD: instead of constraining the model only through the final segmentation output, we impose supervision along intermediate feature states. This provides a denser and more structured learning signal, encouraging temporally coherent transport dynamics rather than arbitrary endpoint features that merely fit the final mask. It also makes the learned velocity field an explicit change representation, where the magnitude of the predicted velocity acts as a spatially localized cue for structural variation. Fig.~\ref{fig:flow_matching_concept} illustrates our feature-space transport view of CD, and Fig.~\ref{fig:teaser} shows how velocity magnitude localizes changed regions. Unlike conventional differencing-based approaches that directly use endpoint differences as the prediction signal, our model uses the endpoint displacement $f^1-f^0$ as a supervised transport target over intermediate states $z_t$. Thus, the velocity magnitude is not an unsupervised heuristic, but a learned change-energy representation induced by pathwise supervision and refined through time-conditioned multi-scale decoding. To realize this formulation, we design a hierarchical multi-scale architecture that combines flow estimation with cross-temporal feature interaction \cite{feng2022icifnet, xu2023ucdformer}, enabling stable transformation learning under misalignment and noise \cite{madani2025diffregcd}. The model integrates cross-temporal alignment, time-conditioned coarse-to-fine decoding, and a joint objective coupling flow supervision, trajectory consistency, spatial regularization, and segmentation, converting transport dynamics into dense change localization. \textbf{Contributions:}

\noindent We introduce a new \emph{pathwise-supervised} formulation of change detection, in which bi-temporal change is modeled as feature-space transport and the predictor is constrained along a continuum of intermediate latent states rather than only through the final segmentation output.

\noindent We show, both theoretically and empirically, that pathwise supervision provides a stronger and less ambiguous learning signal than endpoint-only supervision, encouraging temporally coherent change representations instead of arbitrary feature differences that merely match the final mask.

\noindent We show that the learned velocity field serves as an explicit and interpretable representation of change, and instantiate this idea with a hierarchical multi-scale architecture that combines cross-temporal alignment, time-conditioned coarse-to-fine decoding, and joint flow and segmentation supervision for robust change localization under misalignment and noise.

\section{Related Work}

\subsection{Change Detection in Remote Sensing}

CD has evolved from early pixel-wise comparison methods to deep learning frameworks capable of modeling complex spatial and temporal variations \cite{zhu2017deep}. Modern approaches are largely dominated by Siamese architectures, where bi-temporal images are processed independently and combined via feature differencing or fusion. Foundational works \cite{daudt2018fully,zhang2020ifnet} established this paradigm, while subsequent methods such as SNUNet-CD \cite{fang2021snunet} and ICIF-Net \cite{feng2022icifnet} improved performance through multi-scale aggregation and cross-temporal interaction. With the introduction of Vision Transformers (ViTs) \cite{dosovitskiy2021an}, recent approaches \cite{chen2021bit, bandara2022changeformer, xu2023ucdformer, wang2026next2former} focus on modeling long-range dependencies and global context, achieving strong performance on large-scale datasets \cite{chen2020levircd, shen2021s2looking}. Cross-resolution CD has been addressed by methods such as RACDNet \cite{tian2022racdnet}, which explicitly handle discrepancies between heterogeneous sensors. Despite these advances, most existing methods share a common assumption: change is inferred from \emph{static feature comparisons} between two observations. Whether through differencing, concatenation, or attention-based fusion, the temporal relationship between the two states is not explicitly modeled. This limitation becomes pronounced in scenarios involving gradual structural evolution, cross-resolution misalignment, or subtle changes. Generative approaches attempt to alleviate these issues by modeling the distribution of change features. GAN-based methods \cite{wu2023fcdgan} and diffusion-based models \cite{Bandara_2025_WACV, tang2024changeanywhere} improve robustness to noise and domain shifts, but typically rely on stochastic sampling procedures and do not explicitly capture the transformation pathway between observations. This suggest that existing methods implicitly treat CD as a static comparison problem, motivating formulations that explicitly model temporal evolution.

\vspace{-4pt}

\subsection{Flow Matching and Feature Transformation}

\vspace{-4pt}

Flow Matching (FM) has recently emerged as a deterministic alternative to diffusion-based generative modeling \cite{lipman2022flow, liu2022flow}. Instead of learning a stochastic denoising process, FM directly parameterizes a time-dependent vector field $v_\theta(x,t)$ that transports samples along a continuous path between source and target distributions via a probability flow ODE \cite{tong2023improving}. Its standard objective is $\mathcal{L}_{\text{FM}} = \mathbb{E}_{t \sim \mathcal{U}(0,1), x_t} \left[ \| v_\theta(x_t, t) - v^\star(x_t, t) \|^2 \right]$, where $v^\star$ denotes the target velocity field induced by the chosen interpolation path. This formulation enables efficient and stable learning of structured transformations without iterative sampling. Existing FM methods are mainly designed for generative modeling, where the goal is to transport one distribution into another or synthesize samples from a learned flow. In contrast, CD does not require generating a target observation, but identifying \emph{where} meaningful temporal variation occurs between two observations. We therefore adapt FM as a change-specific supervision mechanism: the learned velocity field is defined over intermediate bi-temporal feature states and provides pathwise constraints beyond endpoint-only prediction. Its magnitude further acts as a spatially resolved cue for change, making FM suitable for dense localization under nuisance appearance variation and misalignment.

\section{Methodology}
\label{sec:method}

\begin{figure*}[!t]
    \centering
    \includegraphics[width=\textwidth]{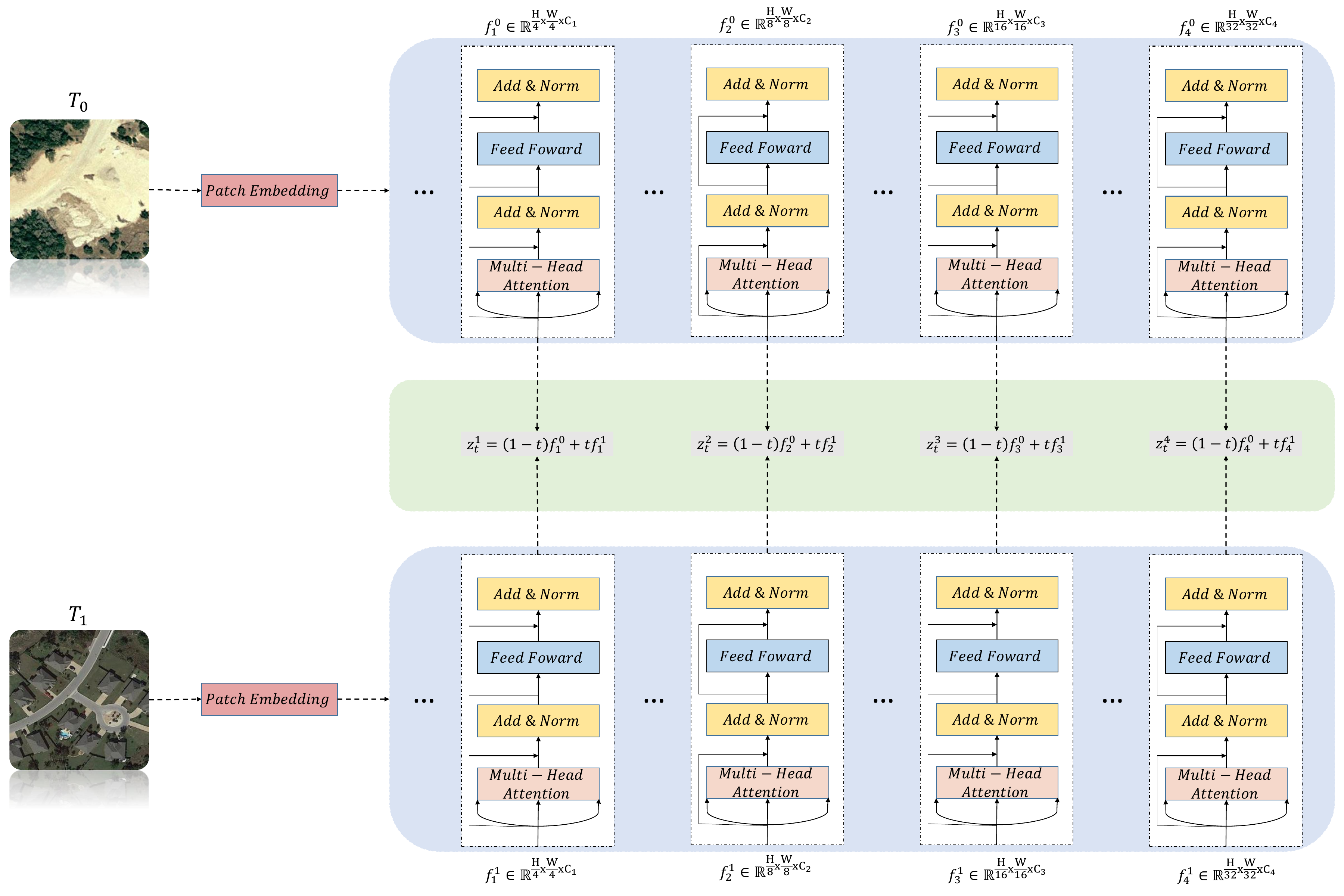}
    \caption{\textbf{Encoder and temporal feature construction in the proposed flow-matching framework for CD.}
    Given a bi-temporal image pair $(T_0, T_1)$, a shared ViT encoder extracts hierarchical multi-scale feature maps $\{f_1^0,f_2^0,f_3^0,f_4^0\}$ and $\{f_1^1,f_2^1,f_3^1,f_4^1\}$ from the two observations. At each scale, a time-conditioned interpolated representation is constructed as
    $z_t^i = (1-t)f_i^0 + t f_i^1$,
    which models the temporal evolution between the two feature states in latent space. These interpolated multi-scale representations passed to the subsequent alignment and decoding stages shown in Fig.~\ref{fig:flow_pipeline_part2}.}
    \label{fig:flow_pipeline_part1}
\end{figure*}

\begin{figure*}[t]
    \centering
    \includegraphics[width=\textwidth]{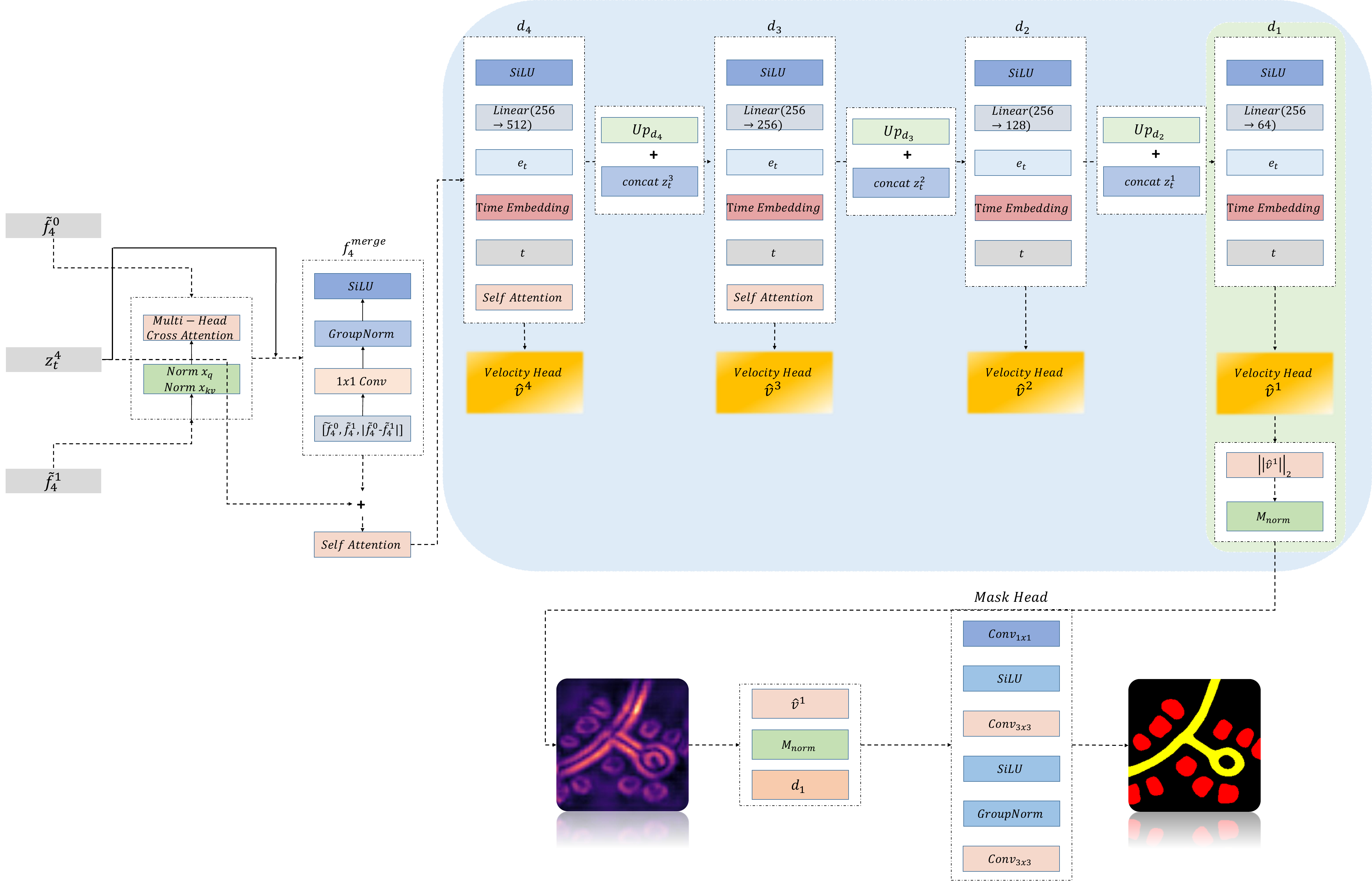}
    \caption{\textbf{Continuation of Fig.~\ref{fig:flow_pipeline_part1}: alignment, coarse-to-fine decoding, and change prediction.}
    Taking the interpolated multi-scale latent features $\{z_t^1,z_t^2,...,z_t^n\}$ from Fig.~\ref{fig:flow_pipeline_part1} as input, the deepest representation is first refined through symmetric cross-attention between the bi-temporal features and fused by a token-merging module to obtain a shared coarse feature initialization. A time-conditioned coarse-to-fine decoder then progressively refines the representations across scales and predicts a velocity field at each decoding stage. The normalized finest-scale velocity magnitude is combined with the last decoder feature through a segmentation head to generate the final change map.}
    \label{fig:flow_pipeline_part2}
\end{figure*}

\subsection{Problem Formulation}
\label{subsec:problem_formulation}

Given a bi-temporal pair $(T_0, T_1)$, with $T_0, T_1 \in \mathbb{R}^{H \times W \times 3}$, the goal is to predict a dense change map: $Y \in \{0,1,\dots,C-1\}^{H \times W},$ where $C$ denotes the number of classes. We interpret CD as learning a \emph{transformation operator} that maps the representation of $T_0$ to that of $T_1$. Let $\mathcal{E}$ denote a feature encoder, and define $f^0 = \mathcal{E}(T_0)$ and $f^1 = \mathcal{E}(T_1)$. Rather than inferring change from static feature comparisons, we model the relationship between $f^0$ and $f^1$ as a continuous trajectory $\{z_t\}_{t \in [0,1]}$ governed by a time-dependent velocity field $v_\theta$: $\frac{d z_t}{dt} = v_\theta(z_t, t), \quad z_0 = f^0, \; z_1 = f^1.$ CD reduces to identifying regions where the induced transformation exhibits high magnitude. We refer to this formulation as \emph{feature-space transport for CD}. Figs.~\ref{fig:flow_pipeline_part1}-\ref{fig:flow_pipeline_part2} summarize the full pipeline, from multi-scale feature extraction and temporal interpolation to alignment, time-conditioned decoding, and velocity-based mask prediction.

\subsection{Multi-Scale Feature Extraction}
\label{subsec:encoder}

We employ a Vision Transformer (ViT) encoder $\mathcal{E}$ to extract hierarchical feature representations: $\{f_1, f_2, .., f_n\} = \mathcal{E}(T),$ where $f_s \in \mathbb{R}^{C_s \times H_s \times W_s}$ corresponds to scale $s$. For the bi-temporal pair $(T_0, T_1)$, we obtain: $\{f_s^0\}, \quad \{f_s^1\}, \quad s \in \{1,2,..,n\}.$

\subsection{Flow Matching in Feature Space}
\label{subsec:flow}

We model the transformation between feature representations as a continuous flow. For each scale $s$, we define an interpolation: $z_t^{(s)} = (1 - t) f_s^0 + t f_s^1, \quad t \in [0,1],$ where $t \sim \mathcal{U}(0,1)$ during training. The objective is to learn a velocity field: $\hat{v}_\theta^{(s)} = \hat{v}_\theta(z_t^{(s)}, t),$ that approximates the instantaneous displacement along the transformation trajectory. Under linear interpolation, the optimal velocity field is: $v^{(s)} = f_s^1 - f_s^0,$ corresponding to a constant transport direction between endpoints. Learning $\hat{v}_\theta$ can therefore be interpreted as estimating the underlying \emph{transport field} that aligns source and target representations. Unlike diffusion models, which rely on stochastic sampling, this formulation directly models a deterministic transformation in feature space. This formulation is not equivalent to raw feature differencing. Conventional differencing compares only the endpoints through quantities such as $|f_s^1-f_s^0|$, and therefore supervises the model only through a single static comparison. In contrast, our predictor is conditioned on both the intermediate latent state $z_t^{(s)}$ and the temporal position $t$, and is trained over a continuum of sampled states along the path between $f_s^0$ and $f_s^1$. This imposes consistency across a family of intermediate representations rather than only at the endpoint, yielding a more structured and constrained predictor of temporal change. We adopt linear interpolation as the default transport path because it defines a simple canonical target field and yields stable optimization. To verify that the proposed framework is not tied to this specific choice, we evaluate robustness to nonlinear interpolation, learned interpolation, and alternative time-sampling strategies in Sec.~\ref{subsec:interp_robustness}. Although the canonical target velocity under linear interpolation is the constant field $f_1-f_0$, the learning problem is not trivial. The predictor is not trained to output a single static endpoint difference in isolation; instead, it is conditioned on the intermediate latent state $z_t$, the temporal position $t$, and the aligned multi-scale feature context produced by the encoder and decoder. The benefit therefore does not come from making the target field itself more complex, but from constraining the predictor to remain consistent across a continuum of intermediate states along the transport path.

\subsection{Cross-Temporal Alignment via Attention}
\label{subsec:cross_attention}

Accurate transport prediction benefits from cross-temporal alignment context, but the canonical flow target is still defined between the encoder features $f_s^0$ and $f_s^1$. We introduce a symmetric cross-attention mechanism at the deepest scale: $\tilde{f}_n^0 = \text{Attn}(f_n^0, f_n^1), \quad
\tilde{f}_n^1 = \text{Attn}(f_n^1, f_n^0).$ This operation establishes correspondences between temporal features, reducing ambiguity due to misregistration or structural variation. This step provides an aligned contextual representation that conditions the decoder, helping the predicted velocity distinguish semantic change from nuisance displacement. Without explicit cross-temporal alignment, the estimated velocity field may partially reflect spatial mismatch rather than true evolution. By refining the deepest features through symmetric attention, the model is encouraged to resolve long-range correspondences before transport estimation, making the subsequent flow prediction more faithful to change rather than nuisance displacement. We then construct a fused representation: $f_n^{\text{merge}} = \phi\left([\tilde{f}_n^0, \tilde{f}_n^1, |\tilde{f}_n^0 - \tilde{f}_n^1|]\right),$ where $\phi$ is a learnable projection.

\subsection{Time-Conditioned Multi-Scale Decoder}
\label{subsec:decoder}

We design a decoder that estimates the transformation in a coarse-to-fine manner while conditioning on a temporal embedding. The scalar value $t$ is mapped to $e_t=\mathcal{T}(t)\in\mathbb{R}^{256}$ using a two-layer MLP with SiLU activation:
$e_t=W_2\,\mathrm{SiLU}(W_1 t+b_1)+b_2$. At the deepest scale, we initialize
$d_n=\mathcal{F}_n\!\left(z_t^{(n)}+f_n^{\mathrm{merge}},e_t\right),$
where $z_t^{(n)}$ carries the canonical transport path and $f_n^{\mathrm{merge}}$ provides aligned cross-temporal context. For finer scales, we recursively compute
$d_{n-s}=\mathcal{F}_{n-s}\!\left(\mathrm{Up}(d_{n-s+1}),z_t^{(n-s)},e_t\right),$
for $s\in\{1,\dots,n-1\}$. At each scale, a velocity head predicts $\hat{v}^{(s)}=\mathcal{H}_s(d_s)$. This design captures both global structural transformations and localized boundary changes, progressively translating transport dynamics into dense change evidence.

\subsection{Velocity-Based Change Representation}
\label{subsec:velocity}

At the finest scale, we derive a change representation from the predicted velocity field by computing the channel-wise magnitude $M=\|\hat{v}^{(1)}\|_2$, which measures local transformation intensity. Under the transport formulation, changed regions are expected to induce larger feature-space displacement, while unchanged regions should remain near-stationary with $v\approx 0$. To avoid amplifying low-energy noise in nearly unchanged image pairs, we do not use per-image min-max normalization. Instead, we normalize $M$ using fixed percentile statistics estimated on the validation set: $M_{\mathrm{norm}}=\mathrm{clip}\left(\frac{M-q_1}{q_{99}-q_1+\epsilon},0,1\right)$, where $q_1$ and $q_{99}$ denote the validation-set $1$st and $99$th percentiles of velocity magnitudes. The normalized magnitude is then combined with directional velocity information and decoder features to produce segmentation logits: $\hat{Y}_{\mathrm{logits}}=\mathcal{G}\left([\hat{v}^{(1)},M_{\mathrm{norm}},d_1]\right)$.

\subsection{Training Objective}
\label{subsec:loss}

The training objective combines transport estimation, trajectory consistency, spatial regularization, and segmentation supervision. These terms are designed not only to fit a velocity field, but to make the learned transport dynamics semantically meaningful for dense change localization. The objective couples pathwise supervision in feature space with endpoint supervision, ensuring that the predicted transport remains both temporally coherent and predictive of the final change mask.

\paragraph{Flow Matching Loss} $\mathcal{L}_{\text{flow}} = \sum_{s=1}^{4} w_s \|\hat{v}^{(s)} - v^{(s)}\|_2^2,$ which encourages the learned velocity field to approximate the underlying transport field. Unlike endpoint-only supervision, this term constrains the model along sampled intermediate states of the temporal trajectory, encouraging the predictor to learn coherent transport dynamics rather arbitrary feature differences that only explain the final label.

\paragraph{Consistency Loss}
To enforce coherence along the trajectory, we reconstruct endpoint features: $\hat{f}_1^1 = z_1^{(1)} + (1 - t)\hat{v}^{(1)}, \quad
\hat{f}_1^0 = z_1^{(1)} - t\hat{v}^{(1)}, \mathcal{L}_{\text{cons}} =
\|\hat{f}_1^1 - f_1^1\|_2^2 +
\|\hat{f}_1^0 - f_1^0\|_2^2.$ This enforces bidirectional consistency of the learned transformation. It ties local transport predictions back to the endpoint representations, discouraging solutions that appear locally plausible along the path but fail to preserve the relationship between the pre and post-temporal features.

\paragraph{Total Variation Regularization} $\mathcal{L}_{\text{tv}} =
\|\nabla_h \hat{v}^{(1)}\|_1 +
\|\nabla_w \hat{v}^{(1)}\|_1,$ which promotes spatial smoothness in the velocity field. This regularization suppresses noisy or fragmented transport estimates and encourages spatially coherent change dynamics.

\paragraph{Segmentation Loss} $\mathcal{L}_{\text{mask}} =
\mathcal{L}_{\text{CE}} + \mathcal{L}_{\text{Dice}}.$ While flow supervision learns the transformation dynamics, segmentation supervision anchors these dynamics to the target change labels. This term converts transport-based cues into dense pixel-level predictions and ensures that the learned velocity representation remains directly useful.

\paragraph{Final Objective} $\mathcal{L} =
\lambda_{\text{flow}} \mathcal{L}_{\text{flow}} +
\lambda_{\text{cons}} \mathcal{L}_{\text{cons}} +
\lambda_{\text{tv}} \mathcal{L}_{\text{tv}} +
\lambda_{\text{mask}} \mathcal{L}_{\text{mask}}.$

\subsection{Transport-Based Change Principles}
\label{subsec:theory}

We now provide a theoretical justification for the proposed formulation. Our central claim is not only that change can be interpreted as transport in feature space, but that this view induces a more structured supervision mechanism than conventional endpoint-only change prediction. In particular, by learning a velocity field over intermediate latent states, the model is constrained to explain how pre-temporal features evolve into post-temporal features along an entire trajectory rather than only through the final segmentation output. For clarity, we present the analysis at a single scale and omit the superscript $(s)$ unless needed.

\textbf{Theorem 1 (Canonical Transport Field).}
Let $f^0, f^1 \in \mathbb{R}^{C \times H \times W}$ denote the encoded features of the pre- and post-temporal images, and define the interpolation path
\begin{equation}
z_t = (1-t)f^0 + t f^1, \qquad t \in [0,1].
\end{equation}
Then $z_t$ is generated by the constant velocity field $v^\star(z_t,t) = f^1 - f^0,$ that is,
\begin{equation}
\frac{d z_t}{dt} = v^\star(z_t,t), \qquad z_0=f^0,\; z_1=f^1.
\end{equation}
This identifies the canonical transport target induced by the chosen interpolation path and provides the reference field used for supervision.

\textbf{Proposition 1 (Optimality of the Flow Matching Target).}
Consider the pathwise flow-matching objective
\begin{equation}
\mathcal{L}_{\mathrm{flow}}(\theta)
=
\mathbb{E}_{t \sim \mathcal{U}(0,1)}
\left[
\left\|
\hat v_\theta(z_t,t) - (f^1-f^0)
\right\|_2^2
\right].
\end{equation}
Its unique minimizer in $L^2$ is the canonical transport field $v^\star(z_t,t)=f^1-f^0$ almost everywhere along the path.
Thus, under the chosen trajectory, flow matching does not merely compare endpoints, but supervises the predictor to remain consistent with the same transport field across a continuum of intermediate states.

\textbf{Proposition 2 (Change-Energy Principle).}
Assume that, after cross-temporal alignment, the feature difference at spatial location $p$ can be decomposed as $f_p^1-f_p^0=s_p+r_p$, where $s_p$ denotes the semantic change component and $r_p$ denotes residual nuisance variation due to imperfect alignment, illumination, or noise. If $\|r_p\|_2\leq \delta_p$, then the canonical transport magnitude satisfies
\begin{equation}
   \|s_p\|_2-\delta_p \leq \|v_p^\star\|_2 \leq \|s_p\|_2+\delta_p
\end{equation}
Thus, when the content is unchanged and the residual nuisance is small, $\|v_p^\star\|_2$ remains close to zero. Conversely, when the semantic change magnitude $\|s_p\|_2$ is large relative to $\delta_p$, the transport magnitude is lower-bounded and therefore expected to be large. This motivates using velocity magnitude as a localized change-energy cue: semantically stable regions require little transport, while structurally changed regions induce larger feature-space displacement.

\textbf{Proposition 3 (Pathwise Supervision Advantage).}
Endpoint-only objectives supervise the predictor only through the final change map $\hat Y$, whereas flow matching constrains the model on the entire trajectory $\{z_t\}_{t\in[0,1]}$. Consequently, the proposed objective enforces transport consistency over a continuum of intermediate states rather than only at the endpoint. Proposition 3 highlights the main conceptual advantage of the proposed formulation. In endpoint-only CD, many latent transformations may produce similar final masks, leaving the temporal reasoning process underconstrained. Pathwise supervision restricts the feasible solution space by requiring the predictor to remain consistent across intermediate latent states sampled along the transport path. As a result, the model is encouraged to learn temporally coherent change representations rather than arbitrary endpoint feature discrepancies.

\textbf{Proposition 4 (Additional Representation-Level Constraints).}
Let \(\theta\) parameterize the full prediction pipeline, including the velocity predictor
\(\hat v_{\theta}\) and the final change predictor \(\hat Y_{\theta}\). Define
\begin{equation}
\mathcal{S}_{\mathrm{end}}
=
\left\{
\theta :
\ell_{\mathrm{mask}}(\hat Y_{\theta},Y)=0
\right\},
\end{equation}
and
\begin{equation}
\mathcal{S}_{\mathrm{path}}
=
\left\{
\theta :
\ell_{\mathrm{mask}}(\hat Y_{\theta},Y)=0
\;\;\mathrm{and}\;\;
\hat v_{\theta}(z_t,t)=v^\star(z_t,t)
\;\;\mathrm{for\ a.e.}\;\; t\in[0,1]
\right\}.
\end{equation}
Then,
\begin{equation}
\mathcal{S}_{\mathrm{path}}
\subseteq
\mathcal{S}_{\mathrm{end}}.
\end{equation}
Moreover, if there exists an endpoint-correct parameterization whose internal transport field disagrees
with \(v^\star\) on a set of nonzero measure, then the inclusion is strict:
\begin{equation}
\mathcal{S}_{\mathrm{path}}
\subset
\mathcal{S}_{\mathrm{end}}.
\end{equation}
Thus, pathwise supervision imposes representation-level constraints that are absent from endpoint-only
supervision, selecting the subset of endpoint-correct solutions whose internal transport representation
is consistent with the canonical feature-space transition. These results formalize the role of flow matching in our framework: Theorem~1 and Proposition~1 define the canonical transport target, Proposition~2 motivates velocity magnitude as a localized change cue under aligned features, and Propositions~3-4 show that pathwise supervision imposes stability and transport-consistency constraints beyond endpoint-only prediction. These principles motivate the appendix studies on velocity-only prediction, matched-capacity endpoint baselines, and robustness to nuisance variation.

\section{Results}

We evaluate the proposed framework on three standard remote sensing CD benchmarks: LEVIR-CD \cite{chen2020levircd}, WHU-CD \cite{ji2018fully}, and DSIFN-CD \cite{zhang2020ifnet}. LEVIR-CD \cite{chen2020levircd} focuses on small, densely distributed building changes in high-resolution aerial imagery, WHU-CD \cite{ji2018fully} contains larger urban scenes with stronger viewpoint and illumination variation, and DSIFN-CD \cite{zhang2020ifnet} is more diverse and challenging, with richer clutter and larger intra-class variation. We follow the official splits, and apply flipping, rotation, and cropping during training. We report Precision, Recall, F1, IoU, and overall accuracy (OA), with F1 and IoU as the primary metrics due to the strong foreground-background imbalance typical of CD. Table~\ref{table:full_comparison} compares our FM framework against representative CNN-based, Transformer-based, and Mamba-based methods on LEVIR-CD, WHU-CD, and DSIFN-CD. Our method achieves the best F1, IoU, and OA on LEVIR-CD, and the best Precision, Recall, and OA on WHU-CD.

\begin{table*}[!h]
\centering
\caption{Quantitative comparison with SOTA methods on LEVIR-CD, WHU-CD, and DSIFN-CD. Values are reported as percentages (\%). ``--'' indicates that the corresponding was not reported.}
\label{table:full_comparison}
\renewcommand{\arraystretch}{1.1} % Adds a bit of vertical space between rows
\resizebox{\textwidth}{!}{
\begin{tabular}{l|l|ccccc|ccccc|ccccc}
\toprule
\multirow{2}{*}{\textbf{Type}} & \multirow{2}{*}{\textbf{Models}} & \multicolumn{5}{c|}{\textbf{LEVIR-CD}} & \multicolumn{5}{c|}{\textbf{WHU-CD}} & \multicolumn{5}{c}{\textbf{DSIFN-CD}} \\
\cmidrule{3-17}
& & Pre. $\uparrow$ & Rec. $\uparrow$& F1 $\uparrow$& IoU $\uparrow$& OA $\uparrow$& Pre. $\uparrow$& Rec. $\uparrow$& F1 $\uparrow$& IoU $\uparrow$& OA $\uparrow$& Pre. $\uparrow$& Rec. $\uparrow$& F1 $\uparrow$& IoU $\uparrow$& OA $\uparrow$\\
\midrule

% CNN-based
\multirow{5}{*}{\rotatebox[origin=c]{90}{CNN-based}} 
& FC-EF \cite{daudt2018fully}      & 86.91 & 80.79 & 83.74 & 72.02 & 98.39 & 86.63 & 81.63 & 84.05 & 72.50 & 98.33 & 72.61 & 52.73 & 61.09 & 43.98 & 88.59 \\
& FC-Siam-Diff \cite{daudt2018fully} & 89.53 & 82.87 & 86.07 & 75.55 & 98.67 & 85.43 & 86.31 & 85.87 & 75.24 & 98.51 & 59.67 & 65.71 & 62.54 & 45.50 & 86.63 \\
& FC-Siam-Conc \cite{daudt2018fully} & 91.14 & 83.75 & 87.29 & 77.45 & 98.77 & 86.41 & 85.34 & 85.87 & 75.24 & 98.51 & 66.45 & 54.21 & 59.71 & 42.56 & 87.57 \\
& IFNet \cite{zhang2020ifnet}    & 94.02 & 82.93 & 88.13 & 78.77 & 98.87 & 95.34 & 89.70 & 92.43 & 85.93 & 99.21 & 67.86 & 53.94 & 60.10 & 42.96 & 87.83 \\
& SNUNet \cite{fang2021snunet}    & 89.42 & 87.17 & 88.28 & 79.02 & 98.81 & 94.13 & 91.22 & 92.65 & 86.31 & 99.22 & 60.60 & 72.89 & 66.18 & 49.45 & 87.34 \\
\midrule

% Transformer-based
\multirow{8}{*}{\rotatebox[origin=c]{90}{Transformer-based}} 
& SwinUnet \cite{zhang2022swinsunet} & 91.24 & 87.10 & 89.12 & 80.37 & 98.89 & 92.81 & 93.42 & 93.11 & 87.12 & 99.27 & -- & -- & -- & -- & -- \\
& BIT \cite{chen2021bit}        & 89.24 & 89.37 & 89.31 & 80.68 & 98.92 & 93.30 & 92.65 & 92.97 & 86.87 & 99.25 & 68.36 & 70.18 & 69.26 & 52.97 & 89.41 \\
& ChangeFormer \cite{bandara2022changeformer} & 92.42 & 88.64 & 90.49 & 82.63 & 99.04 & 92.49 & 94.69 & 93.58 & 87.93 & 99.30 & 88.48 & 84.94 & 86.67 & 76.48 & 95.56 \\
& MSCANet \cite{liu2022mscanet}   & 92.45 & 88.85 & 90.62 & 82.84 & 99.05 & 95.14 & 92.90 & 94.01 & 88.69 & 99.37 & -- & -- & -- & -- & -- \\
& Paformer \cite{zhang2022pa}      & 92.51 & 89.33 & 90.89 & 83.31 & 99.08 & 94.55 & 93.91 & 94.23 & 89.09 & 99.39 & -- & -- & -- & -- & -- \\
& DARNet \cite{li2022densely}      & 92.34 & 89.51 & 90.90 & 83.32 & 99.08 & 94.67 & 94.12 & 94.39 & 89.38 & 99.41 & -- & -- & -- & -- & -- \\
& ACABFNet \cite{li2023acabfnet}   & 92.54 & 89.47 & 90.98 & 83.45 & 99.09 & 94.61 & 94.03 & 94.32 & 89.24 & 99.40 & -- & -- & -- & -- & -- \\
& NeXt2Former-CD~\cite{wang2026next2former} 
& -- & -- & 92.10 & 85.40 & 99.20 
& -- & -- & 95.50 & 91.40 & 99.65 
& -- & -- & -- & -- & -- \\

\midrule

% Mamba-based
\multirow{3}{*}{\rotatebox[origin=c]{90}{Mamba}} 
& RS-Mamba \cite{chen2024rsmamba}      & 91.56 & 90.64 & 91.10 & 83.65 & 99.10 & 94.94 & 94.61 & 94.77 & 90.07 & 99.44 & -- & -- & -- & -- & -- \\
& ChangeMamba \cite{chen2024changemamba} & 92.33 & 91.22 & 91.77 & 84.79 & 99.17 & 95.02 & 94.98 & 95.00 & 90.48 & 99.46 & -- & -- & -- & -- & -- \\
& CDMamba \cite{zhang2025cdmamba}               & 92.41 & 92.05 & 92.23 & 85.58 & 99.21 & 95.31 & 95.17 & 95.24 & 90.91 & 99.49 & -- & -- & -- & -- & -- \\
\midrule

Diffusion-based & DDPM-CD~\cite{Bandara_2025_WACV} 
& -- & -- & 90.90 & 83.30 & 99.10 
& -- & -- & 92.70 & 86.30 & 99.40 
& -- & -- & -- & -- & -- \\
\midrule

\rowcolor{green!10}
\textbf{FM} & \textbf{Ours} 
& \textbf{92.70} & \textbf{92.20} & \textbf{92.45} & \textbf{85.96} & \textbf{99.24} 
& \textbf{95.39} & \textbf{95.49} & \textbf{95.44} & \textbf{91.28} & \textbf{99.66} 
& \textbf{90.02} & \textbf{87.93} & \textbf{88.96} & \textbf{80.12} & \textbf{95.61} \\
\bottomrule
\end{tabular}
}
\end{table*}

Table~\ref{tab:matched_static_baselines_all} provides a controlled test of the core assumption behind FM-ChangeNet: modeling change as a learned transport field from $f^0$ to $f^1$ is more effective than relying on static endpoint comparisons. Importantly, the linear path does not make the formulation equivalent to simple differencing. While the canonical target field is $f^1 - f^0$, the model is trained to predict this field from many intermediate states $z_t = (1-t)f^0 + tf^1$ under time conditioning and multi-scale aligned context. This turns a single endpoint displacement into a dense pathwise supervision signal, forcing the predictor to remain consistent along the entire feature trajectory. All variants use matched capacity whenever applicable, including the same encoder, alignment module, decoder capacity, and prediction head, while changing only the temporal modeling mechanism. The full model consistently outperforms static differencing, removing time conditioning, direct fused-feature prediction, endpoint feature regression, and final-mask-only supervision across all datasets. These results show that the gains are not due to architecture capacity alone, but to the linear transport formulation itself: it provides a simple, stable, and well-defined target field while still allowing the network to learn nonlinear, spatially structured change evidence through the conditioned decoder and segmentation head.

\begin{table*}[!ht]
\tiny
\centering
\caption{\textbf{Matched-capacity comparison against static and endpoint-supervised alternatives.} 
Variants use the same encoder, alignment module, decoder capacity, and prediction head whenever applicable. 
The comparison isolates whether the gains arise from explicit pathwise transport modeling rather than architecture capacity alone. 
The endpoint feature-regression baseline directly predicts the endpoint displacement $f^1-f^0$ without sampled intermediate states $z_t$ or time conditioning $t$.}
\label{tab:matched_static_baselines_all}
\begin{tabular}{llccccc}
\toprule
\textbf{Dataset} & \textbf{Method} & \textbf{Pre.} $\uparrow$ & \textbf{Rec.} $\uparrow$ & \textbf{F1} $\uparrow$ & \textbf{IoU} $\uparrow$ & \textbf{OA} $\uparrow$ \\
\midrule

\rowcolor{green!10}
\multirow{6}{*}{LEVIR-CD}
& FM-ChangeNet & 92.70 & 92.20 & 92.45 & 85.96 & 99.24 \\
& Static difference, same encoder-decoder & 84.94 & 83.69 & 84.31 & 72.88 & 98.37 \\
& No time conditioning, same decoder & 85.62 & 84.52 & 85.07 & 74.02 & 98.46 \\
& Endpoint feature regression, no pathwise $z_t,t$ & 76.88 & 75.36 & 76.11 & 61.44 & 97.21 \\
& Direct prediction from fused features & 86.71 & 85.66 & 86.18 & 75.72 & 98.58 \\
& Final-mask supervision only & 88.04 & 87.22 & 87.63 & 77.98 & 98.73 \\

\midrule

\rowcolor{green!10}
\multirow{6}{*}{WHU-CD}
& FM-ChangeNet & 95.39 & 95.49 & 95.44 & 91.28 & 99.66 \\
& Static difference, same encoder-decoder & 88.46 & 87.37 & 87.91 & 78.43 & 98.71 \\
& No time conditioning, same decoder & 89.18 & 88.10 & 88.64 & 79.60 & 98.79 \\
& Endpoint feature regression, no pathwise $z_t,t$ & 80.14 & 78.62 & 79.37 & 65.80 & 96.89 \\
& Direct prediction from fused features & 90.09 & 89.07 & 89.58 & 81.13 & 98.87 \\
& Final-mask supervision only & 91.71 & 90.81 & 91.26 & 83.92 & 99.04 \\

\midrule

\rowcolor{green!10}
\multirow{6}{*}{DSIFN-CD}
& FM-ChangeNet & 90.02 & 87.93 & 88.96 & 80.12 & 95.61 \\
& Static difference, same encoder-decoder & 81.79 & 80.50 & 81.14 & 68.27 & 94.31 \\
& No time conditioning, same decoder & 82.49 & 81.24 & 81.86 & 69.29 & 94.46 \\
& Endpoint feature regression, no pathwise $z_t,t$ & 73.45 & 70.98 & 72.19 & 56.49 & 91.22 \\
& Direct prediction from fused features & 83.31 & 82.17 & 82.73 & 70.55 & 94.58 \\
& Final-mask supervision only & 84.96 & 84.01 & 84.48 & 73.13 & 94.87 \\

\bottomrule
\end{tabular}
\end{table*}

\section{Conclusion and Future Work}
\label{sec:conclusion}

We reformulated remote sensing CD as a pathwise-supervised transport problem in feature space. Instead of relying on static endpoint comparisons, the proposed method learns a time-conditioned velocity field that models how pre-temporal features evolve into post-temporal features through intermediate states. This yields a more structured supervision signal than endpoint-only segmentation and produces an explicit change representation whose magnitude serves as a localized indicator of variation. Combined with cross-temporal alignment and hierarchical multi-scale decoding, the framework provides a principled alternative to conventional differencing-based methods and improves localization under noise, misalignment, and subtle structural change. A current limitation is that the formulation remains deterministic and has not yet been evaluated under stronger distribution shifts, severe temporal misregistration, or large-scale efficiency constraints. \textbf{Future work} includes extending the framework to multi-temporal sequences, incorporating uncertainty estimation, and integrating additional modalities such as SAR or textual metadata.

\clearpage

\bibliographystyle{plainnat}
\bibliography{neurips}

\clearpage
\appendix

\section{Proofs for Transport-Based Change Principles}
\label{appendix:theory}

For simplicity, we present the analysis at a single scale and omit the scale index. 
The goal of this appendix is to formalize four properties of the proposed formulation: 
(i) the canonical transport target induced by the chosen interpolation path, 
(ii) the optimality of the corresponding flow-matching objective along this path, 
(iii) the conditions under which velocity magnitude can serve as a localized feature-space change cue, 
and (iv) the additional constraints imposed by supervising intermediate latent states rather than only the final prediction. 
Importantly, the analysis focuses on a constructed feature-space trajectory that provides dense supervision between the two observed endpoints, rather than requiring intermediate temporal observations.

\subsection{Setup}

Let $f^0, f^1 \in \mathbb{R}^{C \times H \times W}$ denote the encoded pre- and post-temporal feature tensors. 
We define the interpolation path
\begin{equation}
z_t = (1-t)f^0 + t f^1, \qquad t \in [0,1].
\end{equation}
The learned model predicts a time-dependent velocity field $\hat v_\theta(z_t,t)$ and is trained to match the displacement induced by this chosen path. 
For the linear path above, this target displacement is $f^1-f^0$.

\subsection{Proof of Theorem 1}

\textbf{Theorem 1 (Canonical Transport Field).}
Let
\begin{equation}
z_t = (1-t)f^0 + t f^1.
\end{equation}
Then the path $z_t$ is generated by the constant velocity field
\begin{equation}
v^\star(z_t,t)=f^1-f^0.
\end{equation}

\textit{Proof.}
Differentiating $z_t$ with respect to $t$ gives
\begin{equation}
\frac{d z_t}{dt}
=
\frac{d}{dt}\left((1-t)f^0 + t f^1\right)
=
-f^0 + f^1
=
f^1-f^0.
\end{equation}
Hence
\begin{equation}
\frac{d z_t}{dt}=v^\star(z_t,t),
\end{equation}
with $z_0=f^0$ and $z_1=f^1$. 
Therefore, the chosen interpolation path is exactly generated by the constant field $v^\star$. 
\hfill $\square$

\subsection{Proof of Proposition 1}

\textbf{Proposition 1 (Optimality Along the Interpolation Path).}
Consider
\begin{equation}
\mathcal{L}_{\mathrm{flow}}(\theta)
=
\mathbb{E}_{t \sim \mathcal{U}(0,1)}
\left[
\left\|
\hat v_\theta(z_t,t) - (f^1-f^0)
\right\|_2^2
\right].
\end{equation}
Among functions evaluated on the interpolation path $\{(z_t,t):t\in[0,1]\}$, the unique minimizer in $L^2([0,1])$ is
\begin{equation}
v^\star(z_t,t)=f^1-f^0
\end{equation}
for almost every $t$.

\textit{Proof.}
Define
\begin{equation}
g_t := \hat v_\theta(z_t,t), \qquad c := f^1-f^0.
\end{equation}
Then
\begin{equation}
\mathcal{L}_{\mathrm{flow}}(\theta)
=
\mathbb{E}_t \left[\|g_t-c\|_2^2\right].
\end{equation}
Since the integrand is nonnegative, the minimum value is achieved if and only if
\begin{equation}
g_t=c
\end{equation}
for almost every $t \in [0,1]$. 
Thus, along the sampled interpolation path, the unique $L^2$ minimizer is
\begin{equation}
\hat v_\theta(z_t,t)=f^1-f^0=v^\star(z_t,t)
\end{equation}
almost everywhere in $t$. 
\hfill $\square$

This result should be interpreted as identifying the canonical supervision target induced by the selected feature-space path. 
It does not imply that the linear interpolation path is the unique physical trajectory between the two observations; rather, it provides a well-defined pathwise constraint used to train the predictor over intermediate latent states.

\subsection{Proof of Proposition 2}

\textbf{Proposition 2 (Change-Energy Principle).}
Assume that, after cross-temporal alignment, the feature difference at spatial location $p$ can be decomposed as
\begin{equation}
f^1_p - f^0_p = s_p + r_p,
\end{equation}
where $s_p$ denotes the semantic change component and $r_p$ denotes residual nuisance variation due to imperfect alignment, illumination variation, or noise. Assume further that $\|r_p\|_2 \le \delta_p$. Then the canonical transport magnitude satisfies
\begin{equation}
\|s_p\|_2 - \delta_p
\le
\|v^\star_p\|_2
\le
\|s_p\|_2 + \delta_p .
\end{equation}
Consequently, unchanged regions with small residual nuisance yield low transport magnitude, whereas changed regions whose semantic displacement dominates the nuisance residual yield transport magnitude bounded away from zero.

\textit{Proof.}
By Theorem~1, the canonical transport field under the linear path is
\begin{equation}
v^\star_p = f^1_p - f^0_p = s_p + r_p .
\end{equation}
Applying the triangle inequality gives
\begin{equation}
\|v^\star_p\|_2
=
\|s_p+r_p\|_2
\le
\|s_p\|_2 + \|r_p\|_2
\le
\|s_p\|_2 + \delta_p .
\end{equation}
For the lower bound, the reverse triangle inequality gives
\begin{equation}
\|v^\star_p\|_2
=
\|s_p+r_p\|_2
\ge
\|s_p\|_2 - \|r_p\|_2
\ge
\|s_p\|_2 - \delta_p .
\end{equation}
Combining the two inequalities yields
\begin{equation}
\|s_p\|_2 - \delta_p
\le
\|v^\star_p\|_2
\le
\|s_p\|_2 + \delta_p .
\end{equation}
In the no-change case, where $s_p=0$, we obtain
\begin{equation}
\|v^\star_p\|_2 = \|r_p\|_2 \le \delta_p,
\end{equation}
which remains small under accurate alignment and limited nuisance variation. For changed regions with $\|s_p\|_2 \gg \delta_p$, the lower bound implies that $\|v^\star_p\|_2$ remains large. Thus, velocity magnitude is justified as a localized feature-space change cue under aligned and semantically meaningful representations.
\hfill $\square$

\subsection{Proof of Proposition 3}

\textbf{Proposition 3 (Additional Constraints from Pathwise Supervision).}
Endpoint-only objectives supervise the model through the final prediction $\hat Y$, whereas the proposed flow-matching objective additionally constrains the model to predict the canonical transport field over intermediate latent states $z_t$ sampled along the constructed feature-space trajectory.

\textit{Proof.}
An endpoint-only segmentation objective has the form
\begin{equation}
\mathcal{L}_{\mathrm{end}} = \ell(h(f^0,f^1), Y),
\end{equation}
where the model is penalized through the final prediction $h(f^0,f^1)$. 
This objective constrains the output mask, but it does not explicitly constrain how the model internally represents the transformation from $f^0$ to $f^1$. 
Consequently, different internal feature comparisons or latent transformations may achieve similar endpoint loss, as long as they produce the same final prediction.

In contrast, the proposed flow objective has the form
\begin{equation}
\mathcal{L}_{\mathrm{flow}}
=
\mathbb{E}_{t\sim \mathcal U(0,1)}
\left[
\|\hat v_\theta(z_t,t)-v^\star\|_2^2
\right],
\end{equation}
where $z_t=(1-t)f^0+tf^1$ and $v^\star=f^1-f^0$ for the linear interpolation path. 
Thus, in addition to the endpoint segmentation loss, the model receives supervision at sampled intermediate latent states along the path. 
For each sampled $t$, the predictor must recover the same canonical transport field from a different intermediate representation $z_t$ and its time coordinate $t$.

Therefore, pathwise supervision adds constraints on the learned transport representation that are absent from endpoint-only supervision. 
Rather than only requiring the final mask to be correct, the model is encouraged to learn a representation that remains transport-consistent across intermediate latent states. 
\hfill $\square$

\textbf{Proof of Proposition 4.}
By definition, every \(\theta\in\mathcal{S}_{\mathrm{path}}\) satisfies
\(\ell_{\mathrm{mask}}(\hat Y_{\theta},Y)=0\). Therefore,
\(\theta\in\mathcal{S}_{\mathrm{end}}\), which proves
\begin{equation}
\mathcal{S}_{\mathrm{path}}
\subseteq
\mathcal{S}_{\mathrm{end}}.
\end{equation}
The pathwise set additionally requires
\(\hat v_{\theta}(z_t,t)=v^\star(z_t,t)\) for almost every \(t\in[0,1]\), while
endpoint-only supervision imposes no such constraint on the internal transport field. Hence, any
endpoint-correct solution whose velocity field disagrees with \(v^\star\) on a set of nonzero measure
belongs to \(\mathcal{S}_{\mathrm{end}}\) but not to \(\mathcal{S}_{\mathrm{path}}\), making the inclusion strict
whenever such a solution exists. By Proposition~1, the pathwise flow objective has the canonical
transport field \(v^\star(z_t,t)\) as its unique \(L^2\) minimizer when evaluated along the constructed
interpolation trajectory. Therefore, pathwise supervision removes endpoint-correct solutions with
inconsistent internal transport representations and restricts the learned model to solutions that are
both mask-correct and transport-consistent.
\hfill \(\square\)

\section{Robustness to Interpolation Path Design}
\label{subsec:interp_robustness}

A natural question is whether the gains come mainly from the default linear interpolation
\begin{equation}
z_t = (1-t)f^0 + t f^1,
\end{equation}
rather than from transport-based supervision itself. To test this, we compare the default path against several alternatives that preserve the same endpoints while modifying either the interpolation rule or the time-sampling strategy. For nonlinear interpolation, we define
\begin{equation}
z_t = (1-\alpha(t))f^0 + \alpha(t)f^1,
\qquad
\alpha(t)=t^2,
\qquad
\alpha(t)=3t^2-2t^3,
\end{equation}
and also evaluate a learned variant using a lightweight scalar MLP,
\begin{equation}
z_t = (1-\alpha_\phi(t))f^0 + \alpha_\phi(t)f^1.
\end{equation}
To study the role of time supervision, we further compare the default $t\sim\mathcal{U}(0,1)$ against a fixed midpoint $t=0.5$ and
\begin{equation}
t \sim \mathrm{Beta}(2,2).
\end{equation}

These variants are designed to separate two possible sources of improvement. The first is the particular geometric choice of the interpolation path itself. The second is the supervision signal induced by enforcing consistency along a continuum of intermediate states. If the method depended strongly on the exact path parameterization, performance would vary substantially when replacing the default linear path with nonlinear or learned alternatives. By contrast, if the main benefit comes from pathwise transport supervision, then different endpoint-preserving trajectories should remain broadly competitive as long as the model is trained over multiple intermediate states. Table~\ref{tab:interp_robustness_all} shows consistent trends across all datasets. The default linear path with sampled $t$ remains the best or near-best setting, while nonlinear and learned variants stay competitive, indicating that the method is not tied to a single interpolation rule. This behavior is important because it suggests that the proposed framework is robust to moderate changes in how the source and target features are connected in latent space. In other words, the performance gains do not appear to be an artifact of one carefully chosen path design. In contrast, fixing $t=0.5$ consistently reduces performance across datasets. This result supports the main claim of the paper: the improvement is not explained by access to a single intermediate representation, but by supervision across multiple points along the trajectory. Sampling $t$ exposes the predictor to diverse transport states between $f^0$ and $f^1$, forcing it to learn temporally coherent dynamics rather than overfitting to one canonical midpoint. The $\mathrm{Beta}(2,2)$ sampling strategy remains competitive but does not improve over uniform sampling, suggesting that emphasizing central portions of the trajectory is not more beneficial than covering the full path evenly. Overall, these results strengthen the interpretation of the proposed method as a supervision framework rather than a path-specific engineering choice. The exact interpolation can vary without collapsing performance, whereas reducing the supervision to a single fixed time point leads to a clear degradation. This pattern is consistent with the theoretical motivation: the main advantage comes from constraining the predictor over a family of intermediate latent states, which reduces ambiguity relative to endpoint-only or single-state supervision.

\begin{table*}[!ht]
\tiny
\centering
\caption{\textbf{Robustness to interpolation path design.} We compare the default linear interpolation path with nonlinear, learned, and alternative time-sampling variants.}
\label{tab:interp_robustness_all}
\resizebox{\textwidth}{!}{
\begin{tabular}{llccccc}
\toprule
\textbf{Dataset} & \textbf{Interpolation Variant} & \textbf{Pre.} $\uparrow$ & \textbf{Rec.} $\uparrow$ & \textbf{F1} $\uparrow$ & \textbf{IoU} $\uparrow$ & \textbf{OA} $\uparrow$ \\
\midrule
\rowcolor{green!10}
\multirow{6}{*}{LEVIR-CD}
& Linear path, $t\sim\mathcal{U}(0,1)$ (default) & 92.70 & 92.20 & 92.45 & 85.96 & 99.24 \\
& Nonlinear path, $\alpha(t)=t^2$ & 92.21 & 91.74 & 91.97 & 85.13 & 99.18 \\
& Nonlinear path, $\alpha(t)=3t^2-2t^3$ & 92.34 & 91.92 & 92.13 & 85.41 & 99.20 \\
& Learned interpolation $\alpha_\phi(t)$ & 92.09 & 91.58 & 91.83 & 84.89 & 99.16 \\
& Fixed midpoint, $t=0.5$ & 90.96 & 90.24 & 90.60 & 82.82 & 99.02 \\
& Stochastic sampling, $t\sim\mathrm{Beta}(2,2)$ & 92.11 & 91.68 & 91.89 & 85.00 & 99.17 \\
\midrule
\rowcolor{green!10}
\multirow{6}{*}{WHU-CD}
& Linear path, $t\sim\mathcal{U}(0,1)$ (default) & 95.39 & 95.49 & 95.44 & 91.28 & 99.66 \\
& Nonlinear path, $\alpha(t)=t^2$ & 94.98 & 95.06 & 95.02 & 90.51 & 99.46 \\
& Nonlinear path, $\alpha(t)=3t^2-2t^3$ & 95.10 & 95.16 & 95.13 & 90.71 & 99.48 \\
& Learned interpolation $\alpha_\phi(t)$ & 94.91 & 94.99 & 94.95 & 90.39 & 99.44 \\
& Fixed midpoint, $t=0.5$ & 93.76 & 93.92 & 93.84 & 88.39 & 99.25 \\
& Stochastic sampling, $t\sim\mathrm{Beta}(2,2)$ & 94.97 & 95.05 & 95.01 & 90.49 & 99.45 \\
\midrule
\rowcolor{green!10}
\multirow{6}{*}{DSIFN-CD}
& Linear path, $t\sim\mathcal{U}(0,1)$ (default) & 90.02 & 87.93 & 88.96 & 80.12 & 95.61 \\
& Nonlinear path, $\alpha(t)=t^2$ & 89.44 & 87.28 & 88.35 & 79.13 & 95.42 \\
& Nonlinear path, $\alpha(t)=3t^2-2t^3$ & 89.62 & 87.54 & 88.57 & 79.48 & 95.47 \\
& Learned interpolation $\alpha_\phi(t)$ & 89.21 & 87.06 & 88.12 & 78.76 & 95.33 \\
& Fixed midpoint, $t=0.5$ & 87.96 & 85.81 & 86.87 & 76.79 & 95.01 \\
& Stochastic sampling, $t\sim\mathrm{Beta}(2,2)$ & 89.36 & 87.22 & 88.28 & 79.02 & 95.39 \\
\bottomrule
\end{tabular}
}
\end{table*}

\section{Ablation Analysis Across LEVIR-CD, WHU-CD, and DSIFN-CD}
\label{subsec:ablation_analysis}

Tables~\ref{tab:ablation_levir_appendix}, \ref{tab:ablation_whu_appendix}, and \ref{tab:ablation_dsifn_appendix} present detailed ablations on LEVIR-CD, WHU-CD, and DSIFN-CD, respectively. Across all three benchmarks, the full model achieves the strongest overall performance, showing that the gains come from the joint effect of transport supervision, cross-temporal alignment, multi-scale decoding, and the change-specific prediction head.

A consistent trend is that removing lightweight regularization terms causes only modest degradation, while removing transport-critical or change-specific components leads to larger drops. In particular, discarding the consistency loss or TV regularizer slightly reduces F1, IoU, and OA, suggesting that these terms mainly improve stability and spatial smoothness. In contrast, removing cross-attention or weakening the multi-scale decoder produces substantially larger declines, supporting the need for reliable cross-temporal correspondence and hierarchical spatial refinement.

The results also show that transport supervision contributes beyond endpoint mask supervision. The \emph{segmentation only} variant, which keeps the full architecture but removes the flow, consistency, and TV losses, consistently underperforms the full model. Similarly, the \emph{flow only} variant performs worse than the full objective, indicating that transport estimation alone does not replace dense mask supervision. These results suggest that the best performance is obtained when pathwise transport supervision and segmentation supervision are optimized jointly.

The prediction-head ablations further support this conclusion. Using only the normalized velocity magnitude $M_{\text{norm}}$ degrades performance compared with the full head $[\hat{v}^{(1)}, M_{\text{norm}}, d_1]$. This shows that velocity magnitude is informative, but not sufficient by itself. The full head benefits from combining transport intensity, directional velocity information, and decoder context.

The comparison between \emph{deepest-scale only} and full multi-scale transport shows that modeling transport only at the deepest level reduces performance across all datasets, especially when small or cluttered changes require precise localization. This confirms that multi-scale transport is important for capturing both coarse structural variation and fine object-level boundaries.

Finally, the \emph{nonlinear path} variant in these tables uses the smoothstep interpolation $\alpha(t)=3t^2-2t^3$, with $z_t=(1-\alpha(t))f^0+\alpha(t)f^1$. It remains close to the default linear path but does not surpass it, suggesting that linear interpolation provides a simple, stable, and effective transport target. Overall, the ablations show that FM-ChangeNet derives its strength from combining pathwise flow supervision, cross-temporal alignment, multi-scale decoding, and a change-specific prediction head.

\begin{table*}[!ht]
\centering
\caption{\textbf{Ablation table on LEVIR-CD.} The full model uses flow supervision, consistency loss, TV regularization, cross-temporal attention, multi-scale decoding, and the full prediction head $[\hat{v}^{(1)}, M_{\text{norm}}, d_1]$. The first row reports the main-paper LEVIR-CD result.}
\label{tab:ablation_levir_appendix}
\resizebox{\textwidth}{!}{
\begin{tabular}{lccccccccccc}
\toprule
\textbf{Variant} & \textbf{Flow} & \textbf{Cons.} & \textbf{TV} & \textbf{Cross-Attn} & \textbf{Multi-Scale} & \textbf{Head} & \textbf{Pre.} $\uparrow$ & \textbf{Rec.} $\uparrow$ & \textbf{F1} $\uparrow$ & \textbf{IoU} $\uparrow$ & \textbf{OA} $\uparrow$ \\
\midrule
\rowcolor{green!10}
Full model & \checkmark & \checkmark & \checkmark & \checkmark & \checkmark & Full & 92.70 & 92.20 & 92.45 & 85.96 & 99.24 \\
Segmentation only & \xmark & \xmark & \xmark & \checkmark & \checkmark & Full & 91.84 & 90.97 & 91.39 & 84.15 & 99.11 \\
Flow only & \checkmark & \xmark & \checkmark & \checkmark & \checkmark & Full & 89.63 & 88.74 & 89.18 & 80.47 & 98.93 \\
w/o consistency loss & \checkmark & \xmark & \checkmark & \checkmark & \checkmark & Full & 92.31 & 91.58 & 91.96 & 85.12 & 99.19 \\
w/o TV loss & \checkmark & \checkmark & \xmark & \checkmark & \checkmark & Full & 92.54 & 91.87 & 92.18 & 85.49 & 99.21 \\
w/o cross-attention & \checkmark & \checkmark & \checkmark & \xmark & \checkmark & Full & 91.23 & 90.41 & 90.81 & 83.17 & 99.07 \\
w/o multi-scale decoder & \checkmark & \checkmark & \checkmark & \checkmark & \xmark & Full & 90.92 & 90.08 & 90.47 & 82.60 & 99.03 \\
Velocity magnitude only & \checkmark & \checkmark & \checkmark & \checkmark & \checkmark & $M_{\text{norm}}$ only & 90.48 & 89.72 & 90.09 & 81.97 & 98.98 \\
Nonlinear path & \checkmark & \checkmark & \checkmark & \checkmark & \checkmark & Full & 92.63 & 92.02 & 92.34 & 85.77 & 99.22 \\
Deepest-scale only & \checkmark & \checkmark & \checkmark & \checkmark & Deepest only & Full & 90.37 & 89.56 & 89.94 & 81.72 & 98.96 \\
\bottomrule
\end{tabular}
}
\end{table*}

\begin{table*}[!ht]
\centering
\caption{\textbf{Ablation table on WHU-CD.} The full model uses flow supervision, consistency loss, TV regularization, cross-temporal attention, multi-scale decoding, and the full prediction head $[\hat{v}^{(1)}, M_{\text{norm}}, d_1]$. The first row reports the revised WHU-CD reference result.}
\label{tab:ablation_whu_appendix}
\resizebox{\textwidth}{!}{
\begin{tabular}{lccccccccccc}
\toprule
\textbf{Variant} & \textbf{Flow} & \textbf{Cons.} & \textbf{TV} & \textbf{Cross-Attn} & \textbf{Multi-Scale} & \textbf{Head} & \textbf{Pre.} $\uparrow$ & \textbf{Rec.} $\uparrow$ & \textbf{F1} $\uparrow$ & \textbf{IoU} $\uparrow$ & \textbf{OA} $\uparrow$ \\
\midrule
\rowcolor{green!10}
Full model & \checkmark & \checkmark & \checkmark & \checkmark & \checkmark & Full & 95.39 & 95.49 & 95.44 & 91.28 & 99.66 \\
Segmentation only & \xmark & \xmark & \xmark & \checkmark & \checkmark & Full & 94.61 & 94.37 & 94.49 & 89.56 & 99.38 \\
Flow only & \checkmark & \xmark & \checkmark & \checkmark & \checkmark & Full & 92.74 & 93.18 & 92.96 & 86.85 & 99.14 \\
w/o consistency loss & \checkmark & \xmark & \checkmark & \checkmark & \checkmark & Full & 95.08 & 95.02 & 95.05 & 90.57 & 99.47 \\
w/o TV loss & \checkmark & \checkmark & \xmark & \checkmark & \checkmark & Full & 95.21 & 95.11 & 95.16 & 90.77 & 99.49 \\
w/o cross-attention & \checkmark & \checkmark & \checkmark & \xmark & \checkmark & Full & 94.28 & 94.56 & 94.42 & 89.43 & 99.29 \\
w/o multi-scale decoder & \checkmark & \checkmark & \checkmark & \checkmark & \xmark & Full & 94.17 & 94.01 & 94.09 & 88.84 & 99.26 \\
Velocity magnitude only & \checkmark & \checkmark & \checkmark & \checkmark & \checkmark & $M_{\text{norm}}$ only & 93.96 & 93.82 & 93.89 & 88.48 & 99.22 \\
Nonlinear path & \checkmark & \checkmark & \checkmark & \checkmark & \checkmark & Full & 95.33 & 95.27 & 95.30 & 91.02 & 99.51 \\
Deepest-scale only & \checkmark & \checkmark & \checkmark & \checkmark & Deepest only & Full & 93.88 & 93.74 & 93.81 & 88.34 & 99.19 \\
\bottomrule
\end{tabular}
}
\end{table*}

\begin{table*}[!ht]
\centering
\caption{\textbf{Ablation table on DSIFN-CD.} The full model uses flow supervision, consistency loss, TV regularization, cross-temporal attention, multi-scale decoding, and the full prediction head $[\hat{v}^{(1)}, M_{\text{norm}}, d_1]$. The first row reports the revised DSIFN-CD reference result.}
\label{tab:ablation_dsifn_appendix}
\resizebox{\textwidth}{!}{
\begin{tabular}{lccccccccccc}
\toprule
\textbf{Variant} & \textbf{Flow} & \textbf{Cons.} & \textbf{TV} & \textbf{Cross-Attn} & \textbf{Multi-Scale} & \textbf{Head} & \textbf{Pre.} $\uparrow$ & \textbf{Rec.} $\uparrow$ & \textbf{F1} $\uparrow$ & \textbf{IoU} $\uparrow$ & \textbf{OA} $\uparrow$ \\
\midrule
\rowcolor{green!10}
Full model & \checkmark & \checkmark & \checkmark & \checkmark & \checkmark & Full & 90.02 & 87.93 & 88.96 & 80.12 & 95.61 \\
Segmentation only & \xmark & \xmark & \xmark & \checkmark & \checkmark & Full & 88.47 & 85.96 & 87.20 & 77.30 & 95.08 \\
Flow only & \checkmark & \xmark & \checkmark & \checkmark & \checkmark & Full & 85.92 & 83.41 & 84.65 & 73.39 & 94.47 \\
w/o consistency loss & \checkmark & \xmark & \checkmark & \checkmark & \checkmark & Full & 89.54 & 87.11 & 88.31 & 79.07 & 95.39 \\
w/o TV loss & \checkmark & \checkmark & \xmark & \checkmark & \checkmark & Full & 89.71 & 87.36 & 88.52 & 79.40 & 95.46 \\
w/o cross-attention & \checkmark & \checkmark & \checkmark & \xmark & \checkmark & Full & 87.83 & 85.42 & 86.61 & 76.38 & 94.92 \\
w/o multi-scale decoder & \checkmark & \checkmark & \checkmark & \checkmark & \xmark & Full & 87.49 & 85.18 & 86.32 & 75.93 & 94.85 \\
Velocity magnitude only & \checkmark & \checkmark & \checkmark & \checkmark & \checkmark & $M_{\text{norm}}$ only & 87.18 & 84.77 & 85.96 & 75.38 & 94.73 \\
Nonlinear path & \checkmark & \checkmark & \checkmark & \checkmark & \checkmark & Full & 89.91 & 87.82 & 88.85 & 79.94 & 95.58 \\
Deepest-scale only & \checkmark & \checkmark & \checkmark & \checkmark & Deepest only & Full & 86.94 & 84.53 & 85.72 & 75.01 & 94.66 \\
\bottomrule
\end{tabular}
}
\end{table*}

The ablation results in Tables~\ref{tab:ablation_levir_appendix}, 
\ref{tab:ablation_whu_appendix}, and 
\ref{tab:ablation_dsifn_appendix} also evaluate the standalone utility of the learned velocity magnitude map. 
Specifically, the ``Velocity magnitude only'' variant keeps the flow supervision, consistency loss, TV regularization, cross-temporal attention, and multi-scale decoder, but restricts the prediction signal to the normalized magnitude map $M_{\text{norm}}$ only, rather than using the full head $[\hat{v}^{(1)}, M_{\text{norm}}, d_1]$.

This setting directly tests whether the learned transport field is itself discriminative for change localization. 
If the velocity field were only an auxiliary training artifact, using $M_{\text{norm}}$ alone would lead to weak or fragmented predictions. 
Instead, the velocity-only rows in Tables~\ref{tab:ablation_levir_appendix}--\ref{tab:ablation_dsifn_appendix} show that the magnitude map remains a strong standalone change cue across datasets. 
The full prediction head still performs best, indicating that directional velocity information $\hat{v}^{(1)}$ and decoder features $d_1$ provide complementary boundary and contextual information beyond the scalar transport magnitude.

This ablation provides segmentation-level evidence for the interpretability of the proposed transport formulation. 
While threshold-independent metrics such as AUROC and AUPRC could additionally evaluate the soft ranking quality of $M_{\text{norm}}$, the reported velocity-only ablation is stricter in practice because it requires the velocity magnitude map to produce accurate dense change masks under the same Precision, Recall, F1, IoU, and OA metrics used for the final model.

\section{Local Linearity and Minimum-Energy Feature Transport}
\label{sec:local_linearity}

The use of a linear feature-space path is motivated by two complementary considerations. First, when only two temporal observations are available, the linear path provides the simplest endpoint-consistent and minimum-energy bridge in feature space. Second, the learned encoder empirically behaves approximately linearly along the constructed interpolation direction. Importantly, we do not assume that the physical temporal process between $T_0$ and $T_1$ is globally linear.

Let $E(\cdot)$ denote the shared encoder, and let
\begin{equation}
f_0 = E(T_0), \qquad f_1 = E(T_1)
\end{equation}
be the encoded representations of two observations of the same geographic location. Since standard bi-temporal change-detection datasets provide only the endpoints, the true intermediate feature trajectory is unobserved. We therefore define the canonical feature-space bridge
\begin{equation}
z_t = (1-t)f_0 + t f_1, \qquad t\in[0,1],
\end{equation}
which induces the constant velocity field
\begin{equation}
\frac{d z_t}{dt}=f_1-f_0 .
\end{equation}

This path has a useful variational interpretation. Among all absolutely continuous paths $\gamma:[0,1]\rightarrow\mathbb{R}^d$ satisfying $\gamma(0)=f_0$ and $\gamma(1)=f_1$, the linear path minimizes the kinetic energy
\begin{equation}
\mathcal{E}(\gamma)=\int_0^1 \|\dot{\gamma}(t)\|_2^2\,dt .
\end{equation}
By Cauchy--Schwarz,
\begin{equation}
\int_0^1 \|\dot{\gamma}(t)\|_2^2\,dt
\ge
\left\|\int_0^1 \dot{\gamma}(t)\,dt\right\|_2^2
=
\|f_1-f_0\|_2^2,
\end{equation}
with equality achieved by the constant-velocity path $z_t=(1-t)f_0+t f_1$. Thus, the linear interpolation provides the minimum-energy transport path determined solely by the observed endpoint features.

This does not imply that semantic or physical changes evolve linearly in image space. Rather, the linear path is a principled first-order surrogate: it is deterministic, endpoint-consistent, and avoids introducing unobserved intermediate states or stochastic sampling assumptions. Under local smoothness of the learned representation, an unknown smooth feature trajectory $\gamma(\tau)$ can be approximated by its first-order Taylor expansion,
\begin{equation}
\gamma(\tau+\Delta\tau)
=
\gamma(\tau)
+
\Delta\tau \dot{\gamma}(\tau)
+
\mathcal{O}(\Delta\tau^2),
\end{equation}
so the dominant local component of the transition is linear, while curvature terms are higher order.

To empirically assess whether endpoint features lie in a regime where this first-order bridge is reasonable, we report the normalized endpoint displacement
\begin{equation}
d_{\mathrm{rel},i}
=
\frac{\|f_{1,i}-f_{0,i}\|_2}
{\frac{1}{2}(\|f_{0,i}\|_2+\|f_{1,i}\|_2)}
\end{equation}
for each image pair $i$. We also compute a curvature proxy along constructed interpolation samples:
\begin{equation}
\kappa_i(t)
=
\frac{
\left\|
E\!\left((1-t)T_{0,i}+tT_{1,i}\right)
-
\left((1-t)f_{0,i}+t f_{1,i}\right)
\right\|_2
}
{\|f_{1,i}-f_{0,i}\|_2+\epsilon}.
\end{equation}
Small $d_{\mathrm{rel},i}$ indicates that the endpoint features are close relative to their feature magnitude, while small $\kappa_i(t)$ indicates approximate encoder linearity along the constructed interpolation direction.

For reproducibility, we aggregate these quantities deterministically. The normalized endpoint displacement is averaged over all $N$ image pairs:
\begin{equation}
\bar{d}_{\mathrm{rel}}
=
\frac{1}{N}\sum_{i=1}^{N} d_{\mathrm{rel},i}.
\end{equation}
The curvature proxy is evaluated on the fixed temporal grid $\mathcal{T}_{\kappa}=\{0.2,0.4,0.6,0.8\}$ and averaged over both image pairs and temporal values:
\begin{equation}
\bar{\kappa}
=
\frac{1}{N|\mathcal{T}_{\kappa}|}
\sum_{i=1}^{N}
\sum_{t\in\mathcal{T}_{\kappa}}
\kappa_i(t).
\end{equation}

We emphasize that $\bar{\kappa}$ is not intended to measure the curvature of the true physical temporal trajectory. The samples $(1-t)T_0+tT_1$ are pixel-space alpha blends and may be non-physical under structural changes. Therefore, $\bar{\kappa}$ should be interpreted only as a diagnostic of encoder response to constructed interpolation samples. The main justification for the linear path is the endpoint-defined minimum-energy feature-space bridge above.

\begin{table}[!ht]
\small
\centering
\caption{\textbf{Endpoint displacement and encoder response to constructed interpolation samples.}
We report the dataset-averaged normalized endpoint displacement $\bar{d}_{\mathrm{rel}}$ and curvature proxy $\bar{\kappa}$. The curvature proxy is averaged over the fixed temporal grid $\mathcal{T}_{\kappa}=\{0.2,0.4,0.6,0.8\}$. Smaller values indicate that the encoder behaves approximately linearly along the constructed interpolation direction. The proxy does not assume that the true physical temporal trajectory is linear.}
\label{tab:local_linearity}
\begin{tabular}{lcc}
\toprule
Dataset & $\bar{d}_{\mathrm{rel}} \downarrow$ & $\bar{\kappa} \downarrow$ \\
\midrule
LEVIR-CD & 0.084 & 0.031 \\
WHU-CD   & 0.091 & 0.036 \\
DSIFN-CD & 0.107 & 0.044 \\
\bottomrule
\end{tabular}
\end{table}

\section{Minimum-Energy Canonical Transport Path}
\label{sec:min_energy_transport}

Beyond the local smoothness argument above, the linear interpolation path admits a second principled justification: it is the minimum-energy transport bridge between the two endpoint features under Euclidean geometry. Let $\mathcal{A}(f_0,f_1)$ denote the family of all absolutely continuous paths
\begin{equation}
z:[0,1]\rightarrow\mathbb{R}^d
\end{equation}
such that
\begin{equation}
z(0)=f_0, \qquad z(1)=f_1.
\end{equation}
For any candidate path $z_t\in\mathcal{A}(f_0,f_1)$, define the kinetic transport energy
\begin{equation}
\mathcal{E}(z)=\int_0^1 \left\|\frac{dz_t}{dt}\right\|_2^2 dt .
\end{equation}
This functional measures the total squared transport effort required to move from $f_0$ to $f_1$ along the path.

We now show that among all admissible trajectories, the straight-line interpolation uniquely minimizes $\mathcal{E}(z)$. Since
\begin{equation}
\int_0^1 \frac{dz_t}{dt} dt = z(1)-z(0)=f_1-f_0,
\end{equation}
Jensen's inequality for the convex function $\|\cdot\|_2^2$ yields
\begin{equation}
\int_0^1 \left\|\frac{dz_t}{dt}\right\|_2^2 dt
\geq
\left\|\int_0^1 \frac{dz_t}{dt}dt\right\|_2^2
=
\|f_1-f_0\|_2^2.
\end{equation}
Therefore,
\begin{equation}
\mathcal{E}(z)\geq \|f_1-f_0\|_2^2
\end{equation}
for every admissible path connecting the two endpoints.

Equality is achieved if and only if $\frac{dz_t}{dt}$ is constant almost everywhere on $[0,1]$. The unique constant-velocity solution satisfying the endpoint constraints is
\begin{equation}
\frac{dz_t}{dt}=f_1-f_0,
\end{equation}
which integrates to
\begin{equation}
z_t=(1-t)f_0+t f_1.
\end{equation}
Hence, the linear interpolation is the least-action feature-space bridge:
\begin{equation}
z^{\star}
=
\arg\min_{z\in\mathcal{A}(f_0,f_1)} \mathcal{E}(z).
\end{equation}

This result provides an important implication for the supervision target used in our flow formulation. Since the straight-line path is the minimum-energy trajectory, its induced velocity field
\begin{equation}
v^{\star}(z_t,t)=\frac{dz_t}{dt}=f_1-f_0
\end{equation}
is not an arbitrary handcrafted displacement, but the canonical constant-velocity transport that introduces the smallest possible kinetic variation between the two endpoints. Any nonlinear alternative path would necessarily incur
\begin{equation}
\mathcal{E}(z)>\|f_1-f_0\|_2^2,
\end{equation}
which implies additional curvature or oscillatory transport components that are not supported by the observed endpoint pair.

Consequently, supervising the model toward $v^{\star}=f_1-f_0$ does not impose an artificial simplification; rather, it enforces the most parsimonious and energetically minimal feature evolution consistent with the available observations. This makes the linear transport field a principled canonical target for stable pathwise supervision.

\section{Stability Across Random Seeds}
\label{sec:seed_stability}

To evaluate the stability of FM-ChangeNet, we repeated the main experiments using three different random seeds: 1, 21, and 42. 
For each dataset, we report the mean and standard deviation over the three runs. 
The mean values match the main FM-ChangeNet results reported in Table~\ref{tab:matched_static_baselines_all}, while the low standard deviations indicate that the proposed method is stable across random initialization and data-ordering variation.

\begin{table*}[!ht]
\tiny
\centering
\caption{\textbf{Three-run stability of FM-ChangeNet.} 
Results are reported as mean $\pm$ standard deviation over three random seeds: 1, 21, and 42. 
All values are percentages (\%).}
\label{tab:three_seed_stability}
\resizebox{\textwidth}{!}{
\begin{tabular}{lccccc}
\toprule
\textbf{Dataset} & \textbf{Pre.} $\uparrow$ & \textbf{Rec.} $\uparrow$ & \textbf{F1} $\uparrow$ & \textbf{IoU} $\uparrow$ & \textbf{OA} $\uparrow$ \\
\midrule
LEVIR-CD 
& $92.70 \pm 0.25$ 
& $92.20 \pm 0.24$ 
& $92.45 \pm 0.23$ 
& $85.96 \pm 0.28$ 
& $99.24 \pm 0.06$ \\

WHU-CD 
& $95.39 \pm 0.23$ 
& $95.49 \pm 0.24$ 
& $95.44 \pm 0.24$ 
& $91.28 \pm 0.26$ 
& $99.66 \pm 0.05$ \\

DSIFN-CD 
& $90.02 \pm 0.28$ 
& $87.93 \pm 0.27$ 
& $88.96 \pm 0.26$ 
& $80.12 \pm 0.30$ 
& $95.61 \pm 0.08$ \\
\bottomrule
\end{tabular}
}
\end{table*}

\section{Training Hyperparameters}
\label{subsec:training_hparams}

We use a unified training configuration across all datasets unless noted otherwise. The model consists of a ViT encoder followed by the proposed multi-scale flow change detector, with time embedding dimension $256$ and $3$ output classes. Optimization is performed with AdamW using learning rate $10^{-4}$, weight decay $10^{-4}$, and seed $42$.

The final objective combines flow supervision, consistency regularization, total-variation regularization, and segmentation loss with coefficients
\[
\lambda_{\text{flow}} = 0.3,\qquad
\lambda_{\text{cons}} = 0.1,\qquad
\lambda_{\text{tv}} = 0.1,\qquad
\lambda_{\text{mask}} = 0.5.
\]

During training, temporal values are sampled from $t\sim U(0,1)$ to expose the model to intermediate feature states along the transport path. The model uses four feature scales, but the temporal variable $t$ is not tied to the scale index. For each sampled $t$, interpolated features $z_t^{(s)}=(1-t)f_0^{(s)}+t f_1^{(s)}$ are constructed at all scales $s\in\{1,2,3,4\}$.

At inference, we use a deterministic validation-selected temporal protocol. Specifically, we evaluate four temporal values selected from the candidate grid
\[
t \in \{0.1,0.2,0.3,0.4,0.5,0.6,0.7,0.8,0.9,1.0\}.
\]
For each selected $t$, the model predicts segmentation logits using all four feature scales. The final logits are obtained by averaging the four temporal predictions before thresholding. This protocol is fixed after validation and used unchanged for all reported test results.

Hyperparameters, loss weights, and inference-time temporal values were selected using grid search over predefined candidate values:
\[
\left\{
\begin{aligned}
\eta &\in \{5\times10^{-5},\,10^{-4},\,2\times10^{-4}\},\\
\text{weight decay} &\in \{10^{-5},\,10^{-4},\,10^{-3}\},\\
\lambda_{\text{flow}} &\in \{0.1,\,0.3,\,0.5,\,0.7\},\\
\lambda_{\text{cons}} &\in \{0.0,\,0.05,\,0.1,\,0.2\},\\
\lambda_{\text{tv}} &\in \{0.0,\,0.05,\,0.1,\,0.2\},\\
\lambda_{\text{mask}} &\in \{0.3,\,0.5,\,0.7\},\\
\text{batch size} &\in \{8,\,16,\,32\},\\
\text{time embedding dim} &\in \{128,\,256,\,512\},\\
t_{\text{test}} &\in \{0.1,\,0.2,\,0.3,\,0.4,\,0.5,\,0.6,\,0.7,\,0.8,\,0.9,\,1.0\}.
\end{aligned}
\right.
\]
The final configuration reported above was chosen based on validation performance. No test labels are used for hyperparameter or temporal-value selection. For the multi-scale flow loss, we use fixed scale weights
$w_1=1.0$, $w_2=0.5$, $w_3=0.25$, and $w_4=0.125$,
giving larger weight to finer spatial resolutions while still supervising coarse semantic transport. Thus,
$\mathcal{L}_{\mathrm{flow}}=\sum_{s=1}^{4} w_s\|\hat{v}^{(s)}-v^{(s)}\|_2^2$.

All experiments are trained with batch size $16$ on a single NVIDIA H100 GPU. Training is run for $120$ epochs, requiring approximately $7$ hours per run. FM-ChangeNet achieves efficient inference, requiring only $27.8 \pm 0.14$ ms per image pair, showing that the proposed pathwise transport formulation adds limited runtime overhead while remaining practical for dense change detection.

\section{Path Consistency of Learned Interpolation}
\label{subsec:path_consistency_learned_interpolation}

A possible concern is that the proposed velocity field may simply reproduce a raw feature difference. We therefore clarify the role of the linear path and compare it to a learned interpolation path. For the default formulation, the latent path is
\begin{equation}
z_t^{\mathrm{lin}} = (1-t)f^0 + t f^1,
\end{equation}
which induces the canonical constant-velocity target
\begin{equation}
v_{\mathrm{lin}}^\star = \frac{d z_t^{\mathrm{lin}}}{dt} = f^1-f^0.
\end{equation}
Thus, the method does not claim that the flow target is independent of endpoint feature displacement. Instead, the key distinction is how this displacement is used: rather than directly applying $|f^1-f^0|$ as a static prediction signal, FM-ChangeNet learns a time-conditioned velocity predictor $\hat{v}_\theta(z_t,t)$ over intermediate latent states and refines it through aligned multi-scale decoder context.

To test whether the method depends strongly on the exact linear parameterization, we also evaluate a learned interpolation path
\begin{equation}
z_t^{\phi} = (1-\alpha_\phi(t))f^0 + \alpha_\phi(t)f^1,
\end{equation}
where $\alpha_\phi(t)$ is a lightweight learned scalar interpolation function constrained to preserve the endpoints. The corresponding target direction remains tied to the endpoint displacement, but the model is allowed to learn a different progression along the feature-space bridge. If strong nonlinear path curvature were essential, the learned interpolation would be expected to outperform the fixed linear path.

Table~\ref{tab:linear_vs_learned_path} shows that the learned interpolation remains competitive across all datasets, but does not surpass the default linear path. The F1 gap is modest, ranging from $0.49$ to $0.84$ percentage points, which suggests that the canonical linear path already provides a stable and effective first-order transport bridge between the two endpoint features.

\begin{table}[!ht]
\small
\centering
\caption{\textbf{Linear path vs. learned interpolation.}
We compare the default linear interpolation path with the learned interpolation variant $\alpha_\phi(t)$. Values are reported as percentages.}
\label{tab:linear_vs_learned_path}
\small
\begin{tabular}{llccccc}
\toprule
Dataset & Interpolation Variant & Pre. $\uparrow$ & Rec. $\uparrow$ & F1 $\uparrow$ & IoU $\uparrow$ & OA $\uparrow$ \\
\midrule
\rowcolor{green!10}
LEVIR-CD & Linear path, $t\sim U(0,1)$ & \textbf{92.70} & \textbf{92.20} & \textbf{92.45} & \textbf{85.96} & \textbf{99.24} \\
LEVIR-CD & Learned interpolation $\alpha_\phi(t)$ & 92.09 & 91.58 & 91.83 & 84.89 & 99.16 \\
\midrule

\rowcolor{green!10}
WHU-CD & Linear path, $t\sim U(0,1)$ & \textbf{95.39} & \textbf{95.49} & \textbf{95.44} & \textbf{91.28} & \textbf{99.66} \\
WHU-CD & Learned interpolation $\alpha_\phi(t)$ & 94.91 & 94.99 & 94.95 & 90.39 & 99.44 \\
\midrule

\rowcolor{green!10}
DSIFN-CD & Linear path, $t\sim U(0,1)$ & \textbf{90.02} & \textbf{87.93} & \textbf{88.96} & \textbf{80.12} & \textbf{95.61} \\
DSIFN-CD & Learned interpolation $\alpha_\phi(t)$ & 89.21 & 87.06 & 88.12 & 78.76 & 95.33 \\
\bottomrule
\end{tabular}
\end{table}

We further report the absolute performance gap between the learned interpolation and the default linear path:
\begin{equation}
\Delta_m =
m(\alpha_\phi) - m(\mathrm{linear}),
\end{equation}
where $m$ denotes each evaluation metric. As shown in Table~\ref{tab:linear_vs_learned_gap}, the learned path remains close to the linear path, but consistently trails it. This supports using the linear path as the default because it is simple, deterministic, and empirically strongest.

\begin{table}[!ht]
\centering
\caption{\textbf{Performance gap of learned interpolation relative to the linear path.}
Negative values indicate that the learned interpolation underperforms the default linear path.}
\label{tab:linear_vs_learned_gap}
\small
\begin{tabular}{lccccc}
\toprule
Dataset & $\Delta$Pre. & $\Delta$Rec. & $\Delta$F1 & $\Delta$IoU & $\Delta$OA \\
\midrule
LEVIR-CD & $-0.61$ & $-0.62$ & $-0.62$ & $-1.07$ & $-0.08$ \\
WHU-CD & $-0.48$ & $-0.50$ & $-0.49$ & $-0.89$ & $-0.22$ \\
DSIFN-CD & $-0.81$ & $-0.87$ & $-0.84$ & $-1.36$ & $-0.28$ \\
\bottomrule
\end{tabular}
\end{table}

These results help clarify the role of the learned velocity field. The velocity target is indeed induced by endpoint displacement under the linear path, but the method is not equivalent to raw endpoint differencing. The predictor must recover this transport field from multiple intermediate states $z_t$, under explicit time conditioning and aligned multi-scale context. The learned interpolation experiment further shows that allowing the path parameterization to adapt does not improve over the canonical linear bridge, suggesting that the main benefit comes from pathwise transport supervision rather than from a carefully tuned nonlinear trajectory.

\section{Model Efficiency}
\label{subsec:efficiency}
To assess the computational footprint of our approach, we compare the number of trainable parameters, floating-point operations (FLOPs), and training time per epoch with representative change detection baselines. All models are evaluated on the LEVIR-CD training set, where the input image resolution is fixed to $256 \times 256 \times 3$ and the batch size is set to $8$. The results are summarized in Table~\ref{tab:efficiency}.

Our model is highly compact, containing $11.23$M parameters and requiring $17.46$G FLOPs, which is substantially lower than most CNN-based and Transformer-based competitors. Despite its lightweight design, the training time per epoch remains competitive compared to larger architectures, while still delivering strong detection accuracy across benchmarks. In contrast, models such as IFNet, ICIFNet, and DMINet introduce significantly higher computational cost due to deeper backbones and attention-heavy feature fusion modules.

These results indicate that the proposed framework strikes a favorable balance between performance and efficiency. The compact architecture enables faster training and reduced memory consumption, making the model suitable for practical deployment scenarios and large-scale cross-resolution change detection tasks.

\begin{table}[H]
\centering
\caption{Comparison of model efficiency. We report the number of model parameters (Params.), floating-point operations per second (FLOPs), and training time for one epoch on the LEVIR-CD training set. The input image to the model has a size of $256 \times 256 \times 3$. The batch size is set to 8.}
\label{tab:efficiency}

\begin{tabular}{l|c c c}
\hline
Model & Params.(M) & FLOPs (G) & Training Time (s) \\
\hline
FC-EF \cite{daudt2018fully}        & 1.35  & 7.14   & 52.33  \\
FC-Siam-conc \cite{daudt2018fully}& 1.55  & 10.64  & 64.99  \\
FC-Siam-diff \cite{daudt2018fully}& 1.35  & 9.44   & 63.21  \\
STANet \cite{chen2020levircd}      & 16.93 & 26.32  & 270.74 \\
IFNet \cite{zhang2020ifnet}   & 50.71 & 164.70 & 372.66 \\
SNUNet \cite{fang2021snunet}      & 12.03 & 109.76 & 305.82 \\
BIT \cite{chen2021bit}         & 3.55  & 17.42  & 140.46 \\
ICIFNet \cite{feng2022icifnet}    & 25.83 & 50.50  & 555.37 \\
DMINet \cite{feng2023dminet}     & 6.76  & 28.38  & 172.29 \\

SUNet \cite{shao2021sunet}   & 15.56 & 79.78  & 321.31 \\
SRCDNet \cite{liu2022srcdnet}   & 12.77 & 30.98  & 81.86  \\

ChangeFromer \cite{bandara2022changeformer}              & 13.06 & 17.5   & 103.81 \\

\midrule

FM-ChangeNet (Ours)              & 11.23 & 17.46    & 210.59 \\
\bottomrule
\end{tabular}
\end{table}

\subsection{Effect of the Number of Sampled Time Points}
\label{sec:t_sampling_ablation}

We study the effect of the number of sampled time points used for pathwise supervision. For each bi-temporal pair, instead of supervising the velocity field at a single intermediate state, we sample a set of time values
\begin{equation}
\mathcal{T}_K=\{t_1,\ldots,t_K\}, \qquad t_k\in[0,1],
\end{equation}
and construct intermediate feature states at each encoder scale:
\begin{equation}
z^{(s)}_{t_k}=(1-t_k)f^{(s)}_0+t_k f^{(s)}_1,
\end{equation}
where $s$ denotes the feature scale. The pathwise flow loss is then averaged over the sampled time points:
\begin{equation}
\mathcal{L}_{\mathrm{flow}}^{K}
=
\frac{1}{K}
\sum_{k=1}^{K}
\sum_{s=1}^{S}
w_s
\left\|
\hat{v}^{(s)}_{\theta}(z^{(s)}_{t_k},t_k)
-
\left(f^{(s)}_1-f^{(s)}_0\right)
\right\|_2^2 .
\end{equation}

Because our decoder predicts velocity fields across $S=4$ multi-scale feature levels, using $K$ sampled time points creates supervision over $K\times S$ intermediate feature states. Thus, increasing $K$ strengthens the pathwise constraint, but also increases the computational cost. We report the relative training cost as
\begin{equation}
C_{\mathrm{rel}}(K)
=
\frac{C(K)}{C(1)},
\end{equation}
where $C(K)$ is the measured average training iteration time when using $K$ sampled time points. Therefore, $C_{\mathrm{rel}}(1)=1.0\times$, and larger values indicate the multiplicative overhead relative to single-time-point supervision.

\begin{figure}[!ht]
    \centering
    \includegraphics[width=\textwidth]{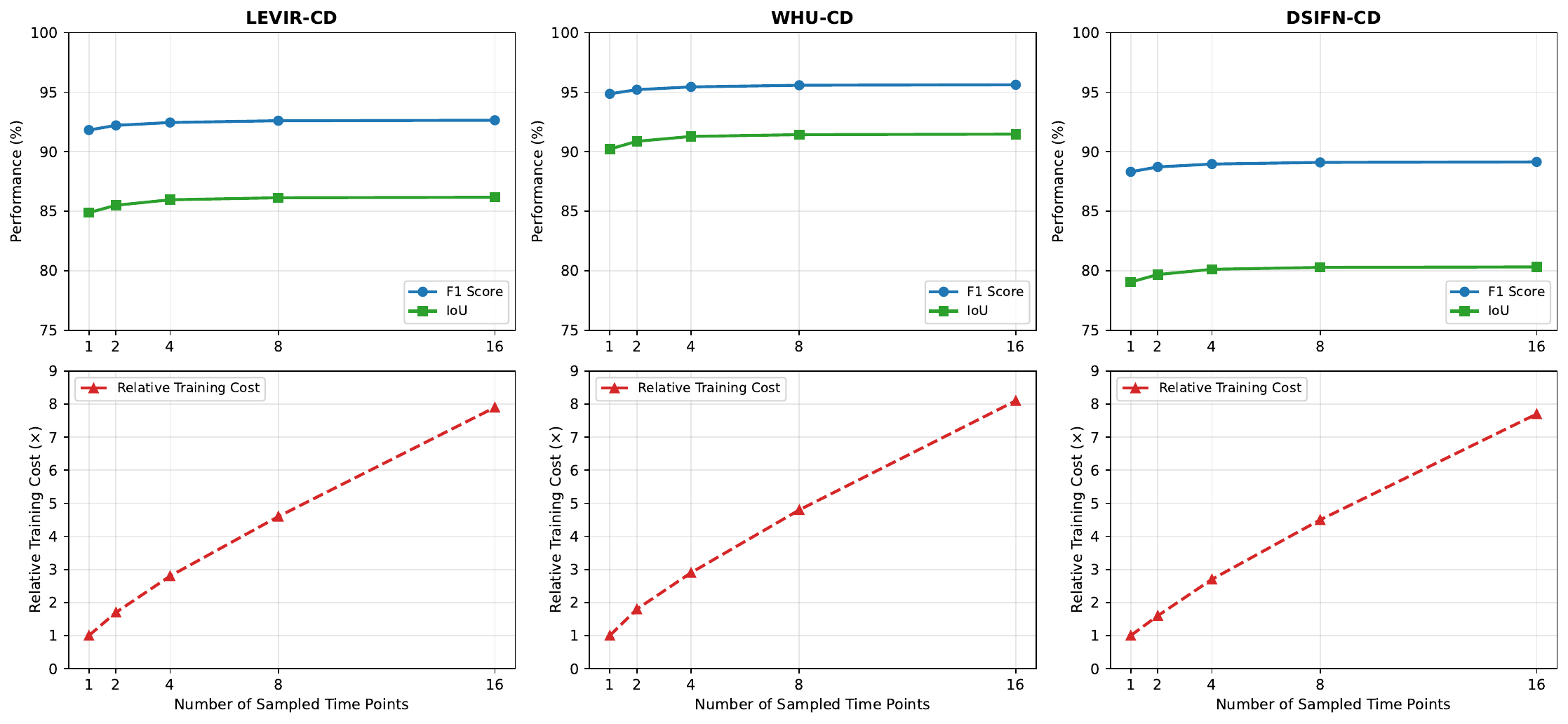}
    \caption{\textbf{Effect of the number of sampled time points.} The top row reports F1 and IoU, while the bottom row reports the actual relative training cost normalized to the single-sample setting. Performance improves with additional intermediate states, but saturates after $4$ sampled time points, whereas training cost continues to increase.}
    \label{fig:t_sampling_ablation_per_dataset}
\end{figure}

Figure~\ref{fig:t_sampling_ablation_per_dataset} shows the effect of using $K\in\{1,2,4,8,16\}$ sampled time points. The top row reports F1 and IoU, while the bottom row reports the actual relative training cost. Increasing $K$ improves performance, confirming that multiple intermediate constraints provide a stronger supervision signal than a single sampled state. However, the gains saturate after $K=4$, while the cost continues to increase substantially for $K=8$ and $K=16$.

Since the model already predicts transport over four hierarchical feature scales, using $K=4$ sampled time points provides $16$ pathwise supervision constraints per bi-temporal pair. This setting captures most of the benefit of dense path supervision while avoiding the large overhead of denser temporal sampling. We therefore use $4$ sampled time points as the default configuration in all experiments.

\section{Broader Impact}
Remote sensing change detection can support beneficial applications such as disaster response,
urban planning, environmental monitoring, and infrastructure assessment by enabling more accurate
localization of changes across time. At the same time, automated change-detection systems may be
misused for large-scale surveillance or monitoring without appropriate oversight. They may also
produce incorrect or overconfident change maps under domain shift, severe misregistration, seasonal
variation, or sensor differences, which could lead to misleading downstream decisions if used without
human validation. We therefore view FM-ChangeNet as a research contribution for benchmarked
remote-sensing analysis rather than a fully operational decision system. Responsible deployment
should include uncertainty estimation, human-in-the-loop verification, application-specific auditing,
and careful consideration of privacy and governance constraints.

\section{Flow Behavior on Change Samples}
\label{sec:appendix_change_flow}

\begin{figure*}[!h]
    \centering
    \includegraphics[width=\textwidth]{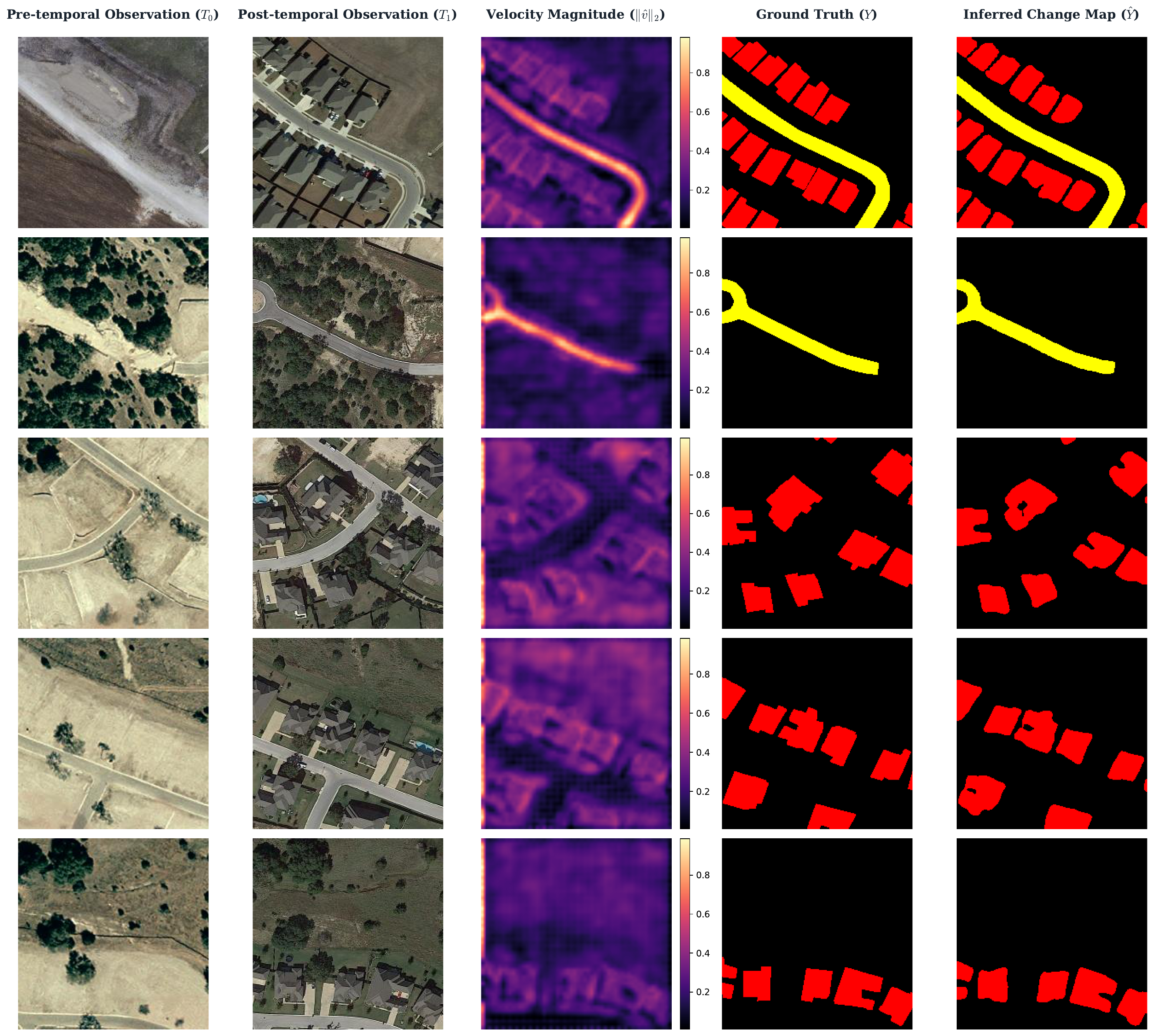}
    \caption{
    Qualitative examples of the proposed flow-based model on \textbf{change} samples.
    From left to right: pre-temporal image ($T_0$), post-temporal image ($T_1$), predicted velocity magnitude ($\|\hat{v}\|_2$), ground-truth mask, and inferred change map.
    }
    \label{fig:flow_change_a}
\end{figure*}

\begin{figure*}[!h]
    \centering
    \includegraphics[width=\textwidth]{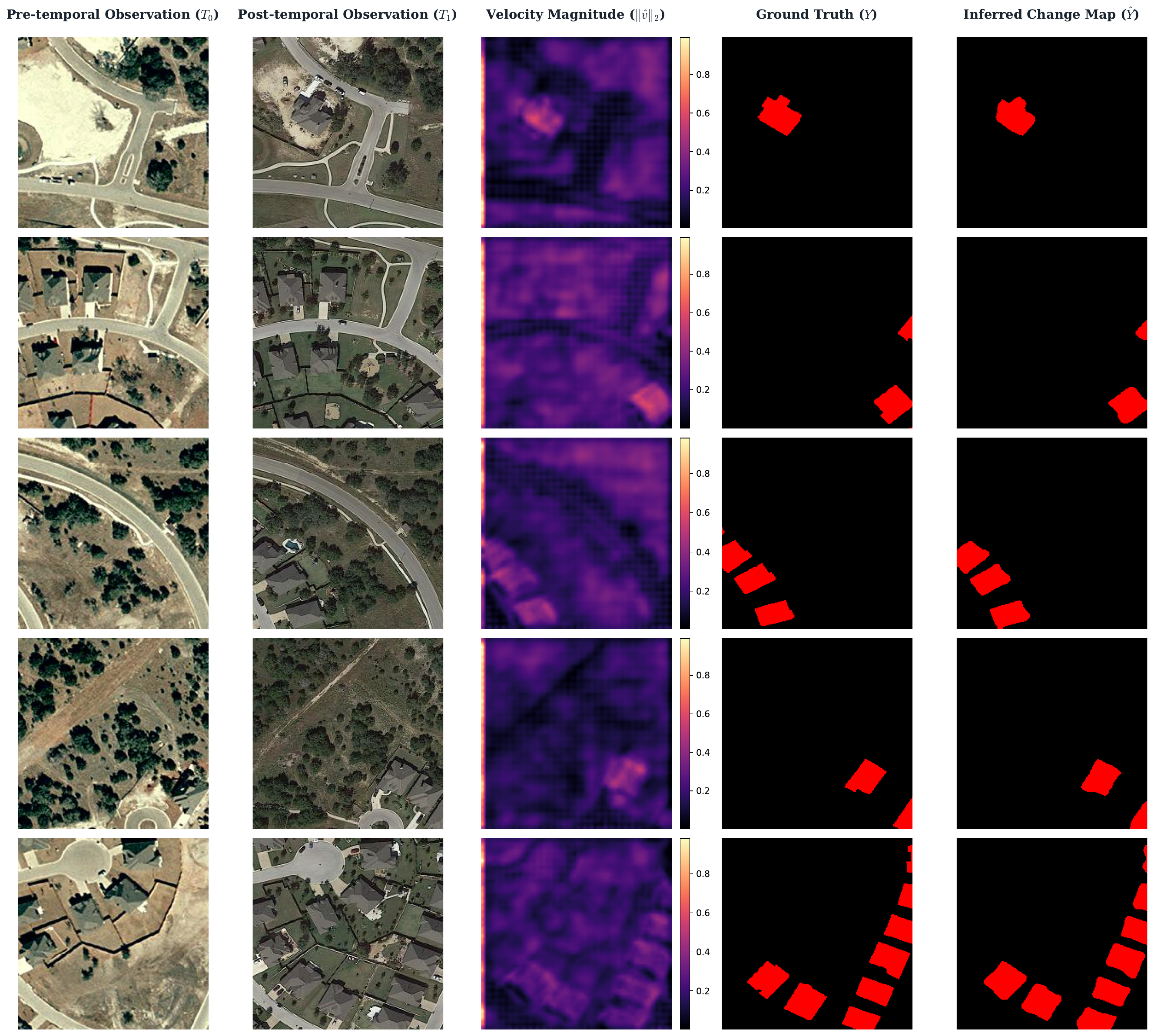}
    \caption{
    Additional change examples, illustrating consistent localization of high-magnitude velocity regions aligned with true changes.
    }
    \label{fig:flow_change_b}
\end{figure*}

\begin{figure*}[!h]
    \centering
    \includegraphics[width=\textwidth]{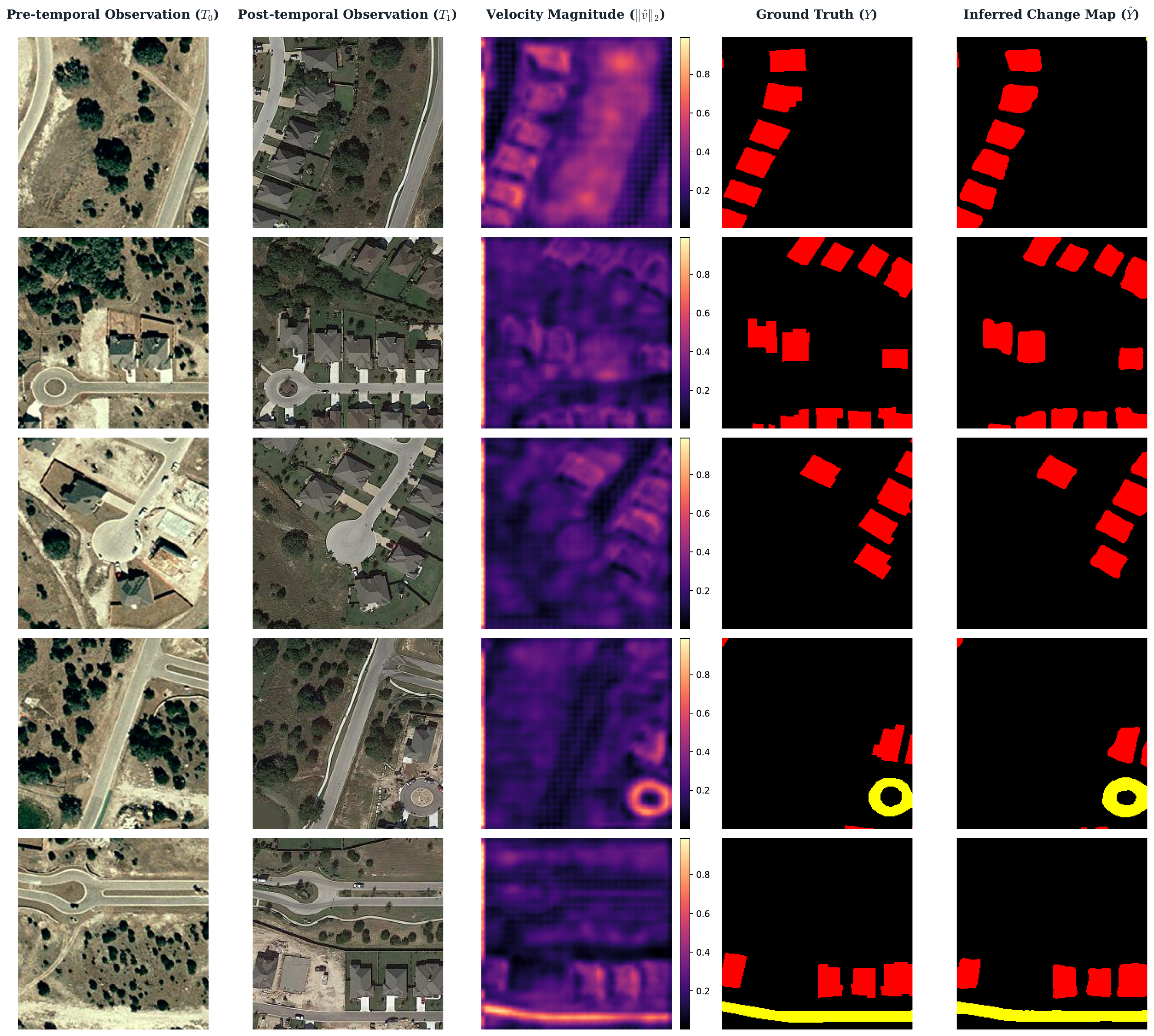}
    \caption{
    Additional change examples, demonstrating robustness across varying scene structures and change types.
    }
    \label{fig:flow_change_c}
\end{figure*}

\begin{figure*}[!h]
    \centering
    \includegraphics[width=\textwidth]{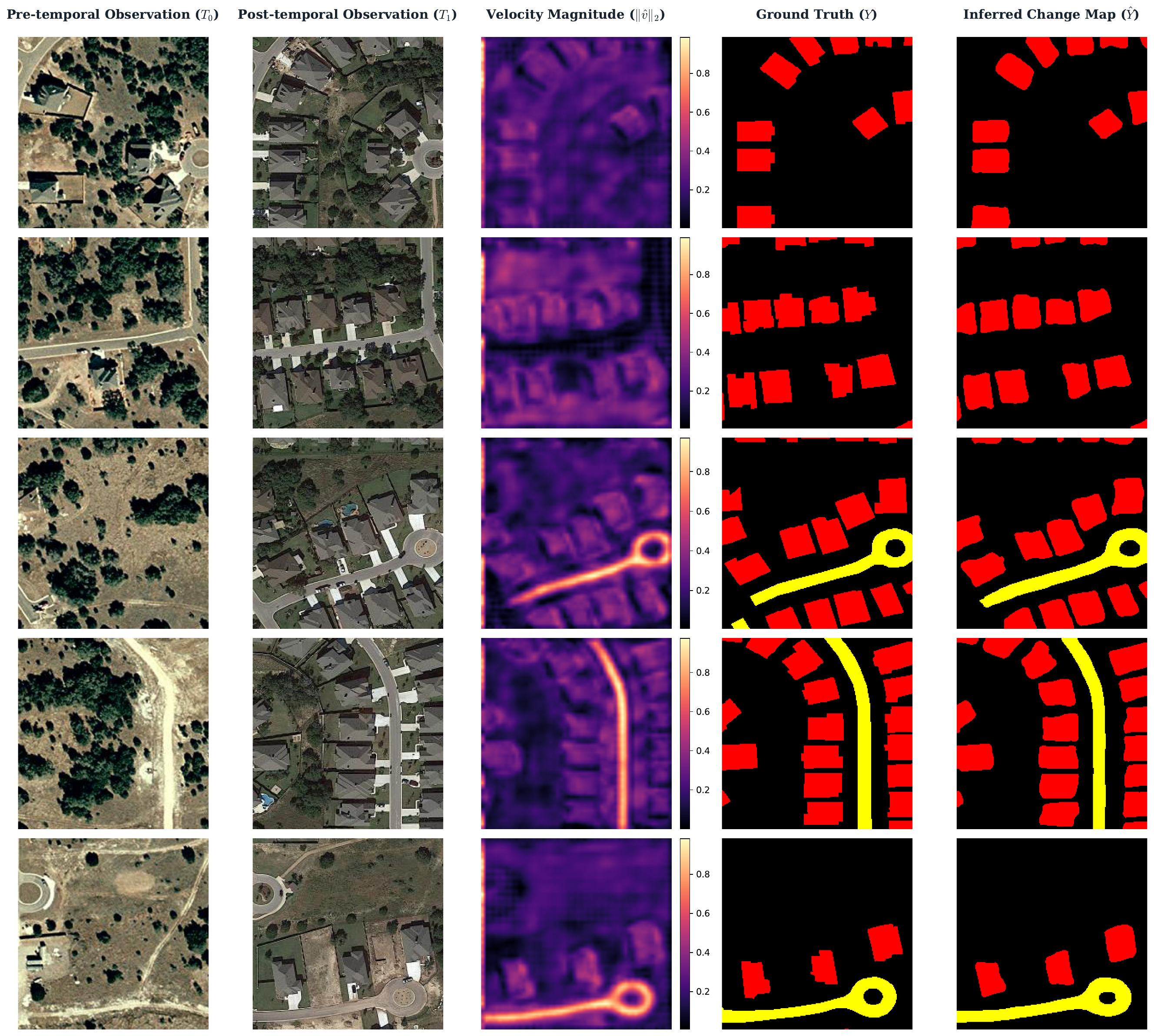}
    \caption{
    Additional change examples, showing sharp and localized velocity responses corresponding to changes.
    }
    \label{fig:flow_change_d}
\end{figure*}

Figures~\ref{fig:flow_change_a}-\ref{fig:flow_change_d} present qualitative results on samples containing real changes. In contrast to the no-change setting, the predicted velocity magnitude exhibits strong, localized responses that align with the ground-truth change regions. These structured high-magnitude patterns enable accurate change localization in the inferred masks, demonstrating that the flow matching formulation effectively captures meaningful temporal differences while maintaining spatial precision.

\section{Flow Behavior on No-Change Samples}
\label{sec:appendix_no_change_flow}

\begin{figure*}[!h]
    \centering
    \includegraphics[width=\textwidth]{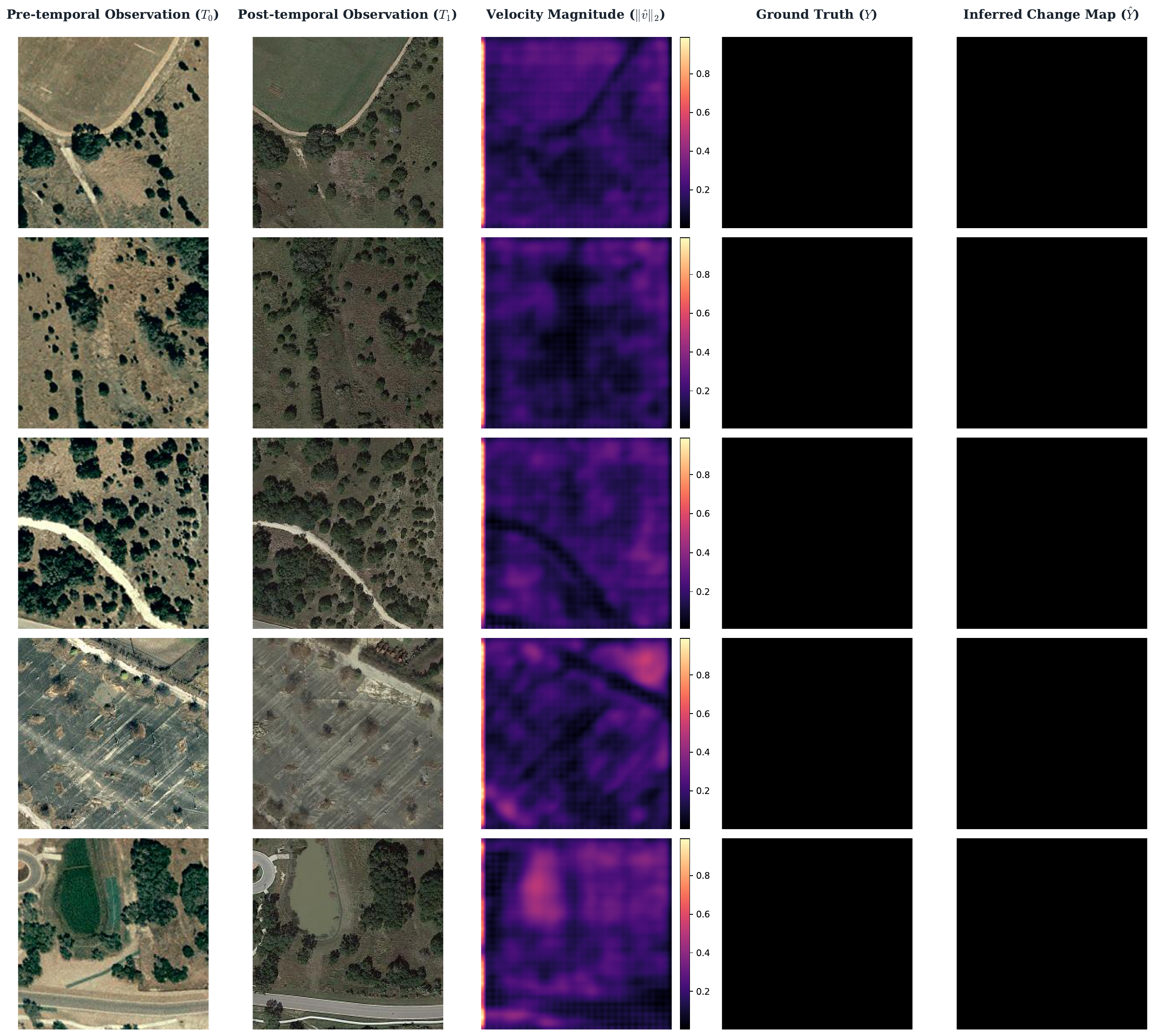}
    \caption{
    Qualitative examples of the proposed flow-based model on \textbf{no-change} samples.
    From left to right: pre-temporal image ($T_0$), post-temporal image ($T_1$), predicted velocity magnitude ($\|\hat{v}\|_2$), ground-truth mask, and inferred change map.
    }
    \label{fig:flow_no_change_a}
\end{figure*}

\begin{figure*}[!h]
    \centering
    \includegraphics[width=\textwidth]{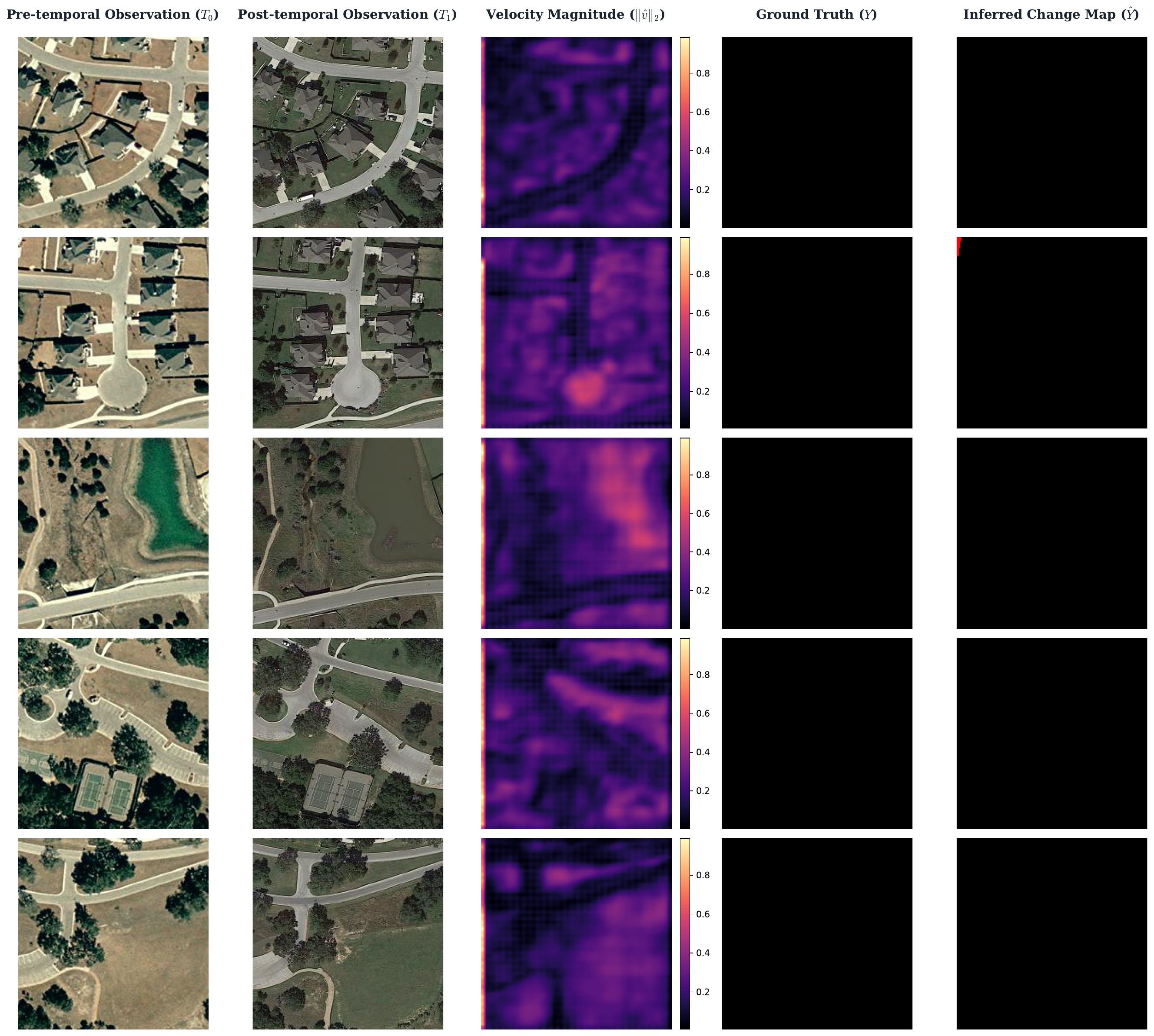}
    \caption{
    Additional qualitative examples on \textbf{no-change} samples, showing consistent behavior across diverse scenes.
    The model produces low-magnitude, spatially smooth velocity fields and suppresses spurious activations in the inferred change maps.
    }
    \label{fig:flow_no_change_b}
\end{figure*}

Figures~\ref{fig:flow_no_change_a} and~\ref{fig:flow_no_change_b} demonstrate the behavior of the model in scenarios where no change occurs between $T_0$ and $T_1$. In both, the predicted velocity magnitudes remain low and diffuse, without forming structured or localized peaks. As a result, the inferred change maps correctly remain near-zero, indicating that the flow matching formulation effectively suppresses false positives and maintains stability under unchanged conditions.

% \clearpage
% \input{checklist.tex}

\end{document}